\documentclass[final]{cvpr}

\usepackage{times}
\usepackage{epsfig}
\usepackage{graphicx}
\usepackage{amsmath}
\usepackage{amssymb}
\DeclareMathOperator*{\argmin}{argmin}
\usepackage{multirow}

\usepackage[ruled,vlined]{algorithm2e}
\makeatletter
\newcommand{\nosemic}{\renewcommand{\@endalgocfline}{\relax}}
\newcommand{\dosemic}{\renewcommand{\@endalgocfline}{\algocf@endline}}
\newcommand{\pushline}{\Indp}

\usepackage{array}
\newcolumntype{M}[1]{>{\centering\arraybackslash}m{#1}}

\usepackage[pagebackref=true,breaklinks=true,colorlinks,bookmarks=false]{hyperref}

\usepackage{paralist}

\def\cvprPaperID{3923} 
\def\confYear{CVPR 2022}

\begin{document}

\title{MuSCLe: A Multi-Strategy Contrastive Learning Framework for Weakly Supervised Semantic Segmentation}

\author{
$\text{Kunhao Yuan}^1 \quad \text{Gerald Schaefer}^1 \quad \text{Yu-Kun Lai}^2$\\
$\text{Yifan Wang}^1 \quad \text{Xiyao Liu}^3 \quad \text{Lin Guan}^1 \quad \text{Hui Fang}^1$\\
{\tt\small $\text{Loughborough University}^1 \quad \text{Cardiff University}^2 \quad \text{Central South University}^3$}
}

\maketitle

\begin{abstract}
Weakly supervised semantic segmentation (WSSS) has gained significant popularity since it relies only on weak labels such as image level annotations rather than pixel level annotations required by supervised semantic segmentation (SSS) methods. Despite drastically reduced annotation costs, typical feature representations learned from WSSS are only representative of some salient parts of objects and less reliable compared to SSS due to the weak guidance during training. In this paper, we propose a novel Multi-Strategy Contrastive Learning (MuSCLe) framework to obtain enhanced feature representations and improve WSSS performance by exploiting similarity and dissimilarity of contrastive sample pairs at image, region, pixel and object boundary levels. Extensive experiments demonstrate the effectiveness of our method and show that MuSCLe outperforms the current state-of-the-art on the widely used PASCAL VOC 2012 dataset.
\end{abstract}

\section{Introduction}
Semantic segmentation is a well-established computer vision task that aims to assign object labels to individual pixels of an image. Various applications, such as autonomous driving, precision agriculture and medical image analysis~\cite{autonomous_driving,precise_agriculture,ronneberger2015unet}, have become possible thanks to recent developments in deep learning (DL)-based semantic segmentation. However, generalising DL models to wider applications is difficult since it requires high-quality pixel level annotations that are costly to obtain. To address this issue, weakly supervised semantic segmentation (WSSS) approaches use inexpensive weak labels, typically image level annotations, to achieve fine-grained semantic segmentation~\cite{zhou2016cam,selvaraju2017grad, kolesnikov2016sec,ahn2019weakly,wang2020seam,chang2020subcat}.

Since the introduction of class activation maps (CAMs)~\cite{zhou2016cam} a lot of effort has focussed on improving DL-based WSSS. One type of approach is to introduce extra cues, such as points~\cite{bearman2016points}, scribbles~\cite{lin2016scribblesup, vernaza2017rw_scribble}, or bounding boxes of objects~\cite{papandreou1502bbox1, dai2015bbox2, khoreva2017bbox3}, to yield stronger constraints to supervise the learning. Another group of methods utilises either global context correlations~\cite{wang2020seam} or local pixel correlations~\cite{ahn2018affinitynet, ahn2019weakly, iclr21pseudo} to enhance image level WSSS.

Despite continuously improved performance of image level WSSS methods, most approaches focus on maximising inter-class variations of feature representations belonging to different classes~\cite{kolesnikov2016sec, ahn2018affinitynet, ahn2019weakly, chen2020weakly}. Consequently, their segmentation results only identify the most salient parts of objects since these are sufficient to optimise their defined loss functions. Although some recent work explores pixel correlations~\cite{wang2020seam} or sub-category clustering in feature space~\cite{chang2020subcat} to enhance object representations, the distinctiveness between contrastive object sample pairs is still under-explored. 

\begin{figure*}[t!]
\centering
\includegraphics[width=0.9\linewidth]{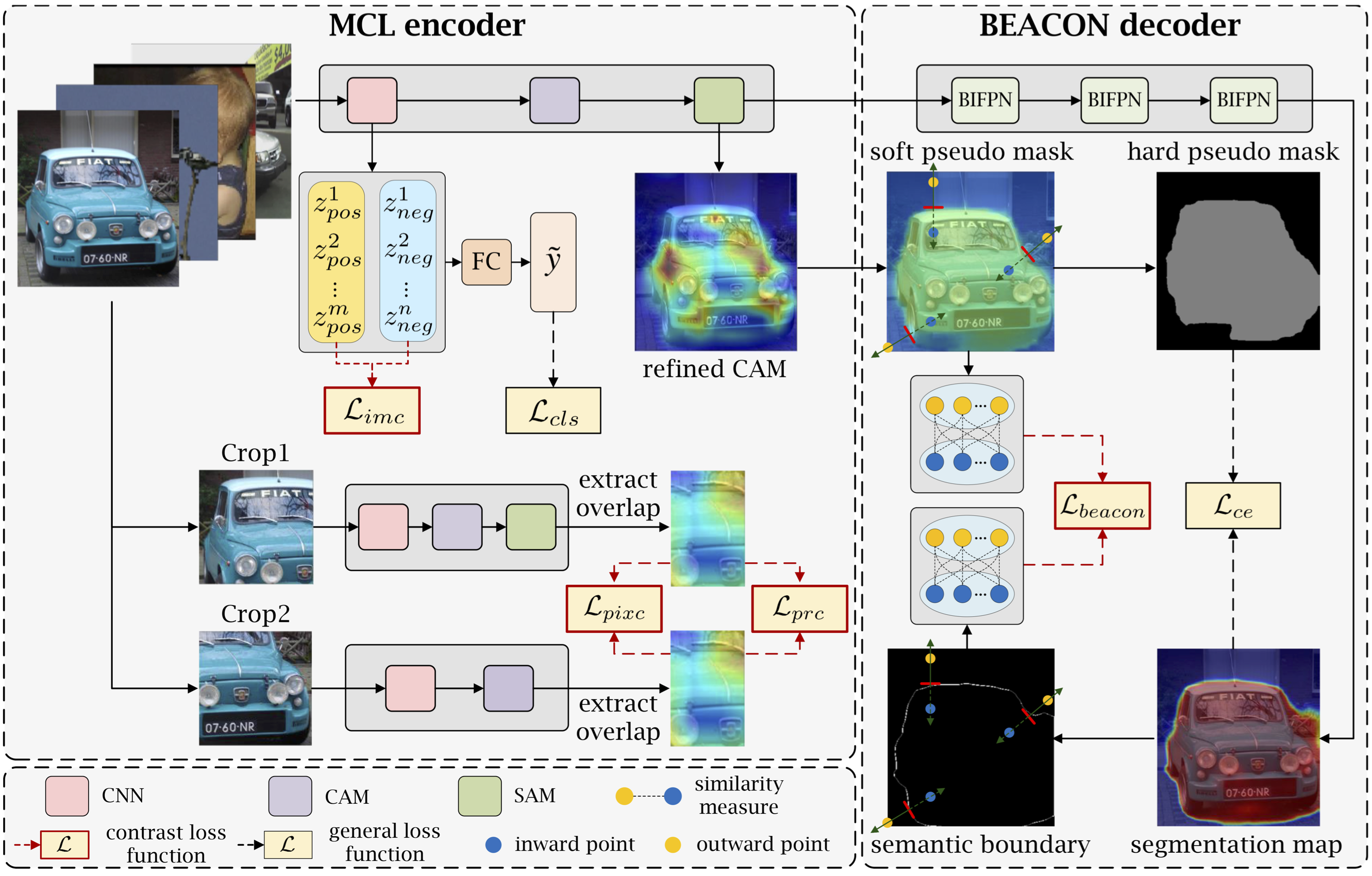}
\caption{Our proposed MuSCLe framework, composed of an MCL encoder and a BEACON decoder, exploits different levels of contrast information to enhance both the feature representation extracted from the encoder and the mapping function of the decoder for better WSSS performance. CAM=class activation map; SAM=spatial attention module; BiFPN=bi-directional feature pyramid network.}
\label{fig:muscle}
\end{figure*}

Inspired by recent success of contrastive learning frameworks~\cite{oord2018cpc, henaff2020cpcv2, chen2020simclr, grill2020byol}, in this paper, we propose a multi-level contrastive learning strategy to further enhance both the feature representation and mapping function of image-level WSSS by embedding contrastive learning metrics at image, region, pixel and object boundary levels. As illustrated in Figure~\ref{fig:muscle}, we use image level contrast between different objects as well as region and pixel level contrast extracted from overlapping regions of the same objects to improve the object feature representation. We further propose a boundary-based contrast extracted from the pseudo labels to enhance our decoder for better segmentation. Overall, the contributions of our Multi-Strategy Contrastive Learning (MuSCLe) framework for WSSS are:
\begin{compactitem}
\item
We propose a multi-contrast learning (MCL) encoder to improve both the  generalisation and distinctiveness of object feature extraction. In addition to the classification loss term used in a typical WSSS method, we explore image level contrast (IMC), pixel level contrast (PIXC), and pairwise regional contrast (PRC) based on overlapping regions of paired randomly cropped object patches to enhance the encoder representation.
\item
We design a novel boundary-based contrastive learning method, Boundary Enhancement viA Contrastive Orientation Navigation (BEACON), to enhance the decoder by learning features across boundaries extracted from pseudo masks, which are derived from our improved activation maps.
\item
Extensive experiments show that MuSCLe outperforms the current state-of-the-art in WSSS on the PASCAL VOC 2012 dataset~\cite{pascal-voc-2012}, and does so with a lighter network architecture (\ie, fewer tuneable parameters).
\end{compactitem}
We will release the code for MuSCLe on GitHub after the blind review process.

\section{Related Work}
\label{sec:rw}
\subsection{Weakly supervised semantic segmentation}
\label{sec:wsss}
WSSS refers to segmenting semantic objects in images at the pixel level when only weak labels are available for training~\cite{kolesnikov2016sec, ahn2019weakly, wang2020seam}. Our focus in this paper is to perform WSSS with only the weakest supervision cue in the form of image labels rather than additional cues such as points~\cite{bearman2016points}, scribbles~\cite{lin2016scribblesup}, or bounding boxes~\cite{khoreva2017bbox3}.

To achieve pixel level semantic segmentation from image level annotations, \cite{kolesnikov2016sec} is one of the pioneering works to use saliency maps generated by a classification network as seeds (\ie, pseudo labels) to guide the training of a segmentation network. To enhance saliency maps, some recent approaches utilise class activation maps as pseudo labels~\cite{zhou2016cam, selvaraju2017grad}. However, early CAM methods only highlight the most distinctive parts of objects, leading to insufficient segmentation performance. More recent approaches explore adversarial erasing~\cite{wei2017erase}, adjacent affinity transformations~\cite{ahn2018affinitynet, ahn2019weakly}, self-supervised attention~\cite{wang2020seam}, sub-category mining~\cite{chang2020subcat}, and boundary exploration~\cite{chen2020weakly} to enhance the quality of pseudo labels.
    
\subsection{Contrastive learning}
\label{sec:cl}
Contrastive learning~\cite{oord2018cpc, bachman2019infomax, henaff2020cpcv2} originates from self-supervised learning~\cite{doersch2015position, noroozi2016jigsaw, pathak2016impaint} and aims to learn generalised feature representations of an image from positive and negative sample pairs. To further explore the pairwise contrast information of image level annotations, \cite{khosla2020supervised} uses image labels to improve the learned features by maximising the distances between paired samples belonging to different classes (\ie, negative samples) while minimising those of the same object pairs (\ie, positive samples).

Introduction of pixel level contrast allows to place constraints on feature maps for better generalisation~\cite{xie2021pixpro}. \cite{wang2020seam} utilises pixel-level contrast from positive samples after geo-transformations to extract so-called equivariant features, while~\cite{liu2020selfemd} employs two feature maps from different Siamese heads as two sets of marginal probability distributions and the earth mover's distance (EMD) to minimise the distance between paired patches in the two sets.

\subsection{Boundary enhancement}
\label{sec:be}
Exploitation of object boundaries is another promising option to enhance WSSS performance. In~\cite{kolesnikov2016sec}, a constrain-to-boundary loss is introduced to align a conditional random field (CRF) with the output of the trained network to support more detailed object segmentation. \cite{ahn2018affinitynet} proposed an affinity network to generate consistent outputs on pixels that share similar semantics by constructing an affinity matrix to enhance object segmentation results, especially at boundaries. The network designed in~\cite{ahn2019weakly} predicts pixel displacements and boundary probabilities to directly obtain an affinity matrix for boundary enhancement, whereas in~\cite{chen2020weakly}, boundary annotations are extracted from an attention-pooling CAM and used to train a boundary exploration net (BENet) to identify object boundaries. These boundary maps form constraints to propagate pixels between salient semantic regions and their corresponding boundaries. Despite its effectiveness, many heuristic parameters need to be set at the training stage to enable BENet to distinguish real object boundaries from low-level edges.

\section{Approach}
In the following, after clarifying the motivation of our work, we then present our novel MuSCLe framework in detail, introducing its multi-contrast learning (MCL) encoder and its Boundary Enhancement viA Contrastive Orientation Navigation (BEACON) decoder.

\subsection{Motivation}
State-of-the-art WSSS methods use enhanced CAMs to generate pseudo labels to provide supervision on an encoder-decoder-based network for semantic image segmentation. Inspired by the recent success of employing contrastive learning to improve feature representations, we design a multi-level contrastive learning framework to enforce multiple constraints to learn more reliable and distinctive feature representations. Paired samples for contrastive learning are extracted at image level, pixel level, regional level and boundary level in order to ensure consistency of features in the same object classes while maximising distances between different object categories. This simple yet effective strategy facilitates the generation of high-quality pseudo labels as well as improves the encoder-decoder network for better segmentation performance.     

\subsection{MCL encoder}
\label{MCL}
The encoder in a WSSS network not only extracts salient feature representations to be used in its decoder but generates pseudo masks to provide additional cues for fine-grained segmentation. As illustrated in Figure~\ref{fig:muscle}, we propose contrastive learning loss terms to build our multi-contrast learning encoder which can generate generalised feature representations and high-quality pseudo masks. 

\vspace{-0.3cm}
\subsubsection{Image level contrast}
\vspace{-0.1cm}
Given a query sample $x_{i}\in{X}$ and its label $y_{i}=(y_{i;1},y_{i;2},...,y_{i;K})$, a $K$-dimensional multi-hot vector to represent the presence of $K$ objects, in each training batch 
 $x_{i}$ and $x_{j}$ form a positive pair if $y_{j}=y_{i}$. Conversely, $x_{i}$ and $x_{j}$ form a negative pair if $y_{j}\cap y_{i}=\varnothing$.
 
We propose a novel way to process contrast pairs in each batch which significantly increases the efficiency of contrastive learning compared to Siamese networks. To measure similarity of both positive and negative pairs, we extract image embeddings by average pooling of the feature maps from the last convolutional layer of the CNN feature extractor and using the dot product to calculate scores.

The image-level contrastive learning loss term we employ is calculated as 
\begin{equation}
\mathcal{L}_{imc}=-\log(\frac{\sum_{Z^{+}}\exp(z_{i}\cdot \Tilde{z_{i}})}{\sum_{Z^{+}}\exp(z_{i}\cdot \Tilde{z_{i}})+\sum_{Z^{-}}\exp(z_{i}\cdot z_{j})}) , 
\label{eq:loss_imc}
\end{equation}
where $z_{i}$ is the vector embedding of the query sample, $\Tilde{z_{i}} \in Z^{+}$ represents each embedding of positive samples and $z_{j} \in Z^{-}$ represents each embedding of negative samples. In contrast to~\cite{henaff2020cpcv2,chen2020simclr}, our $\mathcal{L}_{imc}$ does not rely on augmented views to generate positive samples which significantly reduces memory consumption. Additionally, to alleviate single positive pair bias and to enforce batch-wise attention on positive pairs, we compute the integral of $\exp(z_{i}\cdot \Tilde{z_{i}})$ before taking the logarithm. We show, empirically and theoretically, that this leads to more effective training compared to the loss term from~\cite{khosla2020supervised} in the Supplemental Material.

\vspace{-0.3cm}
\subsubsection{Pixel level contrast}
\vspace{-0.1cm}
Given two random regions cropped from an image, pixel level contrastive learning aims to maximise the feature similarity of pixels in their overlapping region even though their representations are not exactly the same due to their distinct contexts within the receptive field. As highlighted in~\cite{xie2021pixpro}, pixel level contrast imposes pixel-wise feature consistency to enhance the feature representations for its dense-prediction downstream task (\ie, image segmentation in this paper).
    
\begin{figure*}[t!]
\centering
\includegraphics[width=0.85\linewidth]{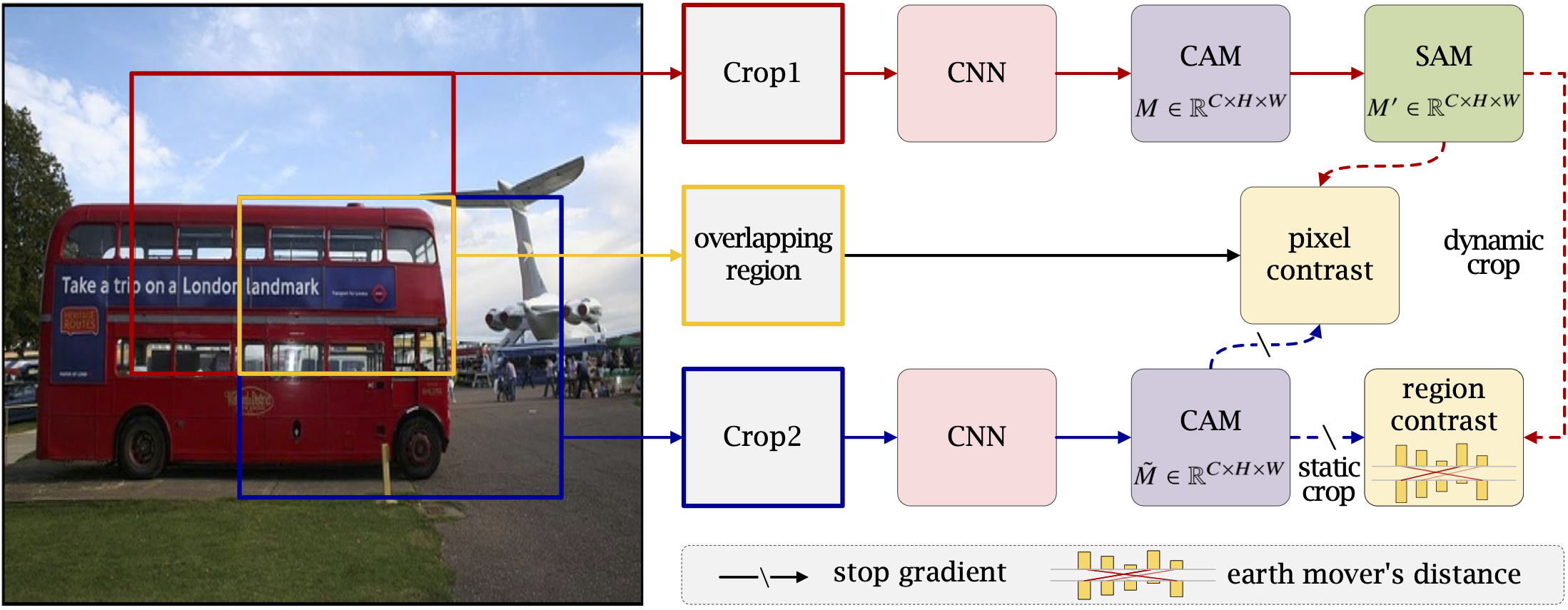}
\caption{Illustration of pixel level contrast and pairwise regional contrast. Blocks of the same colour share the same weights.}
\label{fig:pixcon}
\end{figure*}
As illustrated in Figure~\ref{fig:pixcon}, we obtain pixel level contrastive loss by calculating the similarity between paired pixel-wise features in the overlapping region from two types of feature maps, the original CAM and a spatial attention module (SAM) map \cite{cao2019gcnet, wang2020seam}. 

The SAM utilises a global self-attention mechanism, capable of exploiting long-range contexts, to enhance the CAM, and is obtained as
\begin{equation}
M^{\prime} = SAM(M) = \text{softmax}(g_{1}(M)^T\times g_{2}(M))\times g_{3}(M),
\label{eq:att}
\end{equation}
where $g_{1}(\cdot), g_{2}(\cdot)$, and $g_{3}(\cdot)$ denote individual linear projections and $M$ is the CAM response map. The alignment between CAM and SAM in our contrastive loss further improves the generalisation of the feature representations.

The loss term we employ is
\begin{equation}
\label{eq:pix}
\mathcal{L}_{pixc}=-\frac{1}{HW}\sum_{k=1}^{HW}\cos(u_{k}, \text{sg}(v_{k}^{\prime})),
\end{equation}  
where $u_{k}^{\prime},v_{k}$ are feature vectors from overlapping regions in $M^{\prime}$ and $\tilde{M}$, respectively, $H$ and $W$ are the height and width of the regions and $\text{sg}(\cdot)$ denotes the stop gradient operator which avoids interference of cross-optimisation~\cite{grill2020byol,chen2021simsam}. This design also reduces the computational cost, thus improving the efficiency of our method.

\vspace{-0.3cm}
\subsubsection{Pairwise regional contrast} 
\begin{figure}[t]
\centering
\includegraphics[width=0.9\linewidth]{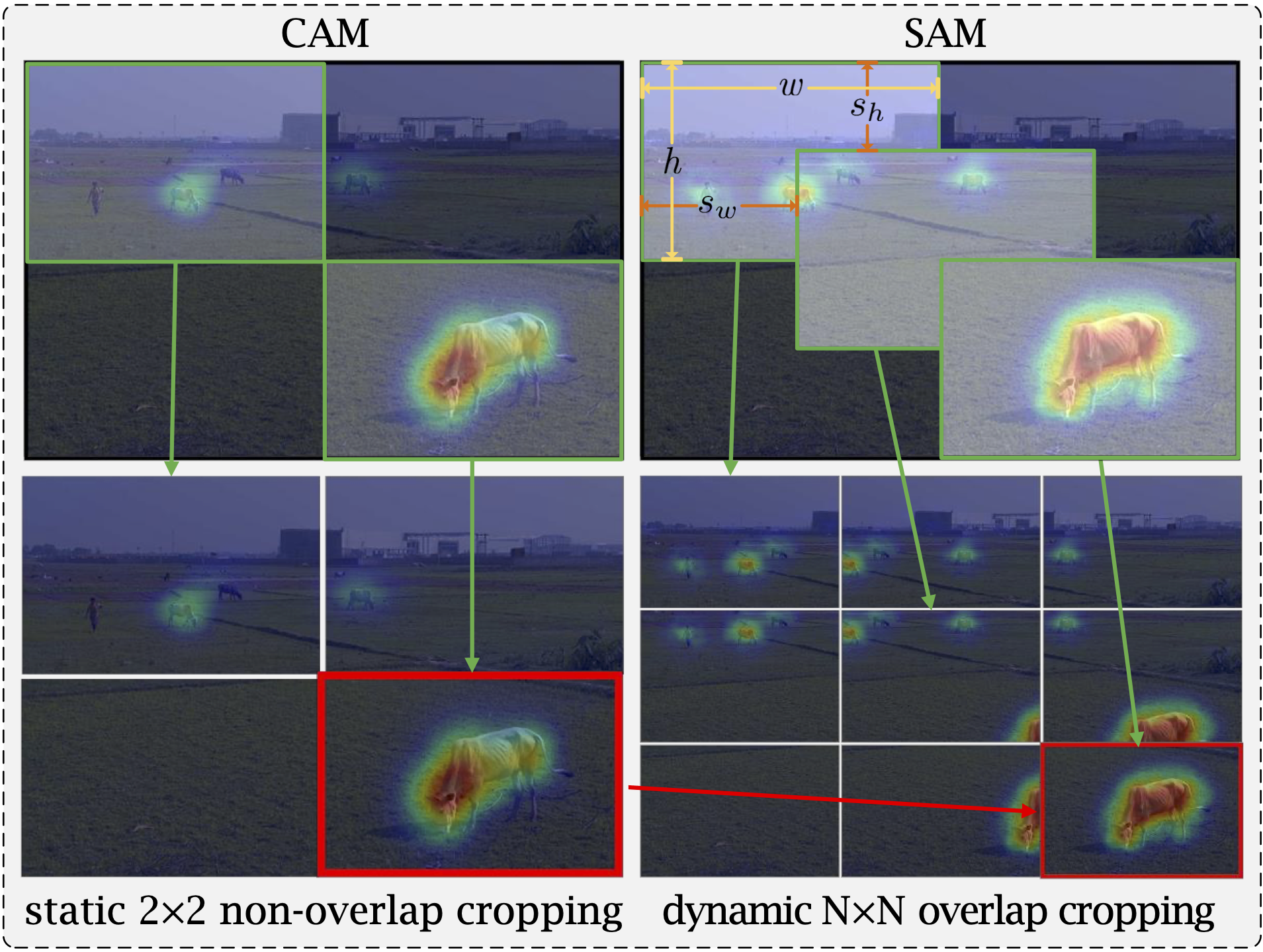}
\caption{Illustration of dynamic cropping and matching.}
\label{fig:dynamic_crop}
\end{figure}

\vspace{-0.1cm}
In addition to image level and pixel level contrastive loss terms, we propose a novel dynamic pairwise regional contrastive loss term to further enhance the scale-invariant characteristics of the extracted features. As illustrated in Figure~\ref{fig:dynamic_crop}, to provide sufficient flexibility while keeping complexity low, we divide the CAM response map into static (\eg $2\times2$) non-overlapping patches. To impose feature consistency from paired objects of different scales, we introduce four parameters, patch width $w$, patch height $h$, and horizontal and vertical sliding strides $s_w$ and $s_h$, to randomly sample regions at different scales from the SAM feature map. We use the earth mover's distance~\cite{cuturi2013sinkhorn, liu2020selfemd} to match and compare the feature maps from paired patches.
    
To obtain a more reliable EMD measure, we aim to avoid the bias introduced from background response maps, and estimate the background activation map as
\begin{equation}
M_{bg}= 1-\max\limits_{1{\le}c{\le}C-1}(M_{c}),
\label{eq:cam_bg}
\end{equation}
with
\begin{equation}
M_{c}=\frac{\exp(M_{c})}{\sum\limits_{1{\le}c{\le}C-1}\exp(M_{c})} ,
\end {equation}
where $M_{c}$ belongs to one of the $C-1$ foreground activation maps and $M_{bg}$ represents the estimated background activation map. The concatenation of all foreground activation maps and the background activation map yields the background-included CAM ($M\in \mathbb{R}^{C\times H\times W}$). We refer the reader to \cite{zhang2020deepemd} for details on marginal weight generation in the EMD computation. 
    
Our pairwise regional contrastive loss term is defined as
\begin{equation}
\mathcal{L}_{prc}=\argmin\limits_{(a,b)}\text{EMD}(p_{a}, \text{sg}(\tilde{p}_{b})),
\label{eq:emd}
\end{equation}
where $p_{a} \in \{p\}_1^{\mathrm{A}}\subset M^{\prime}$ and $\tilde{p}_{b} \in \{\tilde{p}\}_1^{\mathrm{B}} \subset \tilde{M}$ are feature vectors from paired patches generated from the original feature maps $M^{\prime}$ and $\tilde{M}$, respectively. As for pixel level contrastive loss, we use the stop gradient operator $\text{sg}(\cdot)$ to avoid cross-optimisation.

\vspace{-0.3cm}
\subsubsection{Overall loss function}
\vspace{-0.1cm}
In addition to the loss terms introduced above, we use the well-established multi-label multi-class classification loss (\ie, binary cross entropy loss)~\cite{nam2014large}, focal loss~\cite{lin2017focal}, and a pair loss~\cite{li2017pairloss} to address sample imbalance and over-confidence of negative sample issues, and combine them to form a hybrid classification loss (HCL) term
\begin{equation}
\label{eq:hcl}
\mathcal{L}_{hcl}(y, \hat{y}) =  \mathcal{L}_{bce}(y, \hat{y}) + \mathcal{L}_{focal}(y, \hat{y}) + \mathcal{L}_{pair}(y, \hat{y}) ,
\end{equation}
which improves WSSS performance compared to using individual terms.

The overall loss function of our MCL encoder is then defined as
\begin{equation}
\mathcal{L}_{MCL} = \mathcal{L}_{hcl}+\mathcal{L}_{imc}+\mathcal{L}_{pixc}+\mathcal{L}_{prc}.
\end{equation}

\subsection{BEACON decoder}
\label{beacon}

\begin{figure}[b!]
\centering
\renewcommand*{\arraystretch}{0.2}
\setlength{\tabcolsep}{0.05em}
\begin{tabular}{cc}
\includegraphics[width=.47\linewidth, height=.4\linewidth]{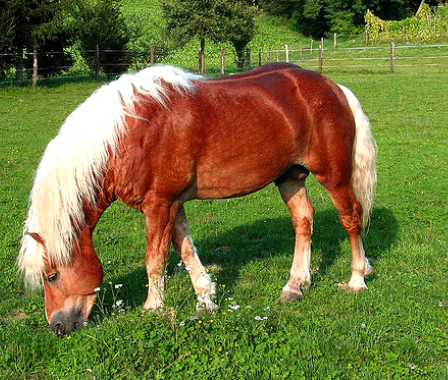} & 
\includegraphics[width=.47\linewidth, height=.4\linewidth]{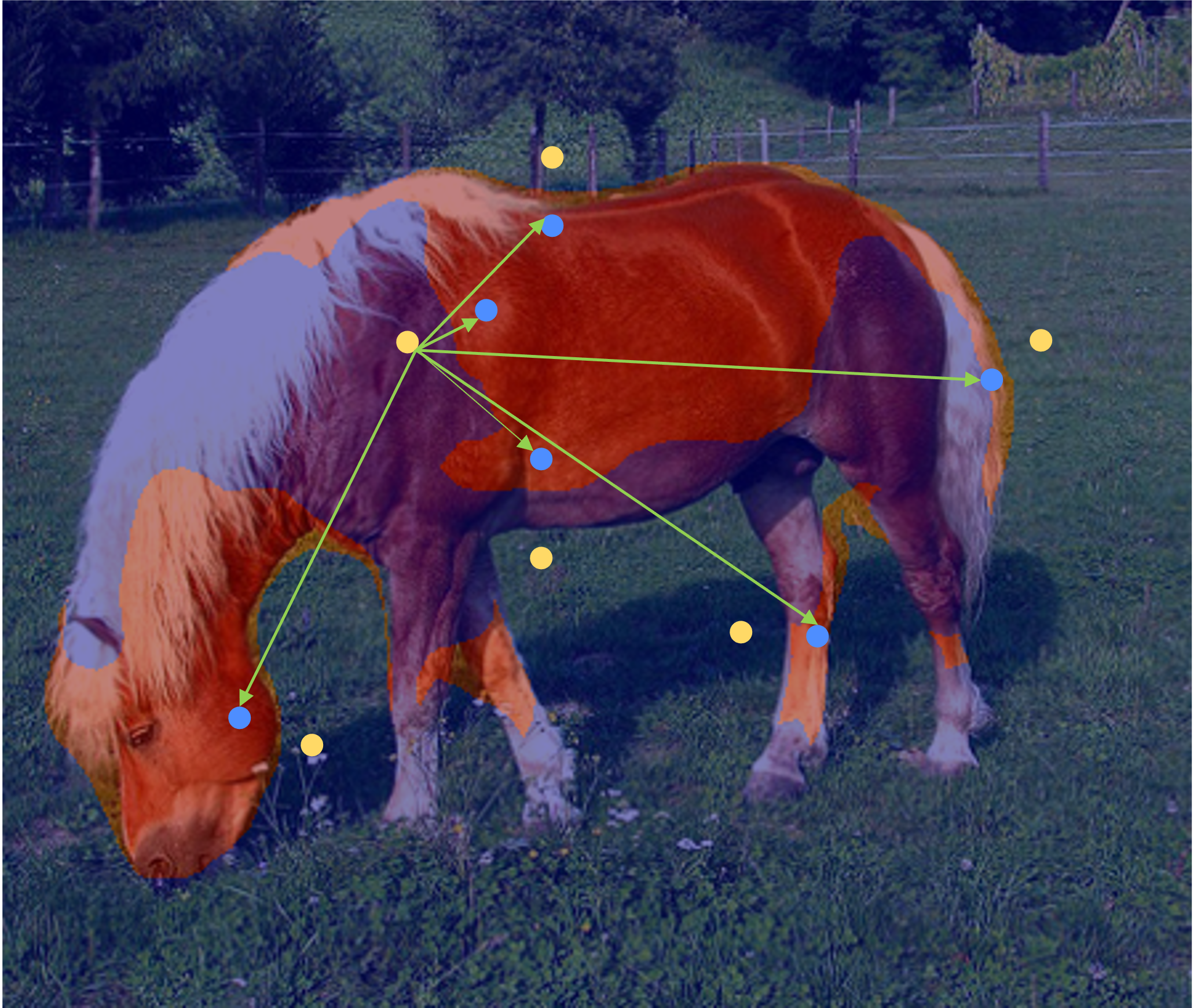} \\
(a) & (b)\\
\includegraphics[width=.47\linewidth, height=.4\linewidth]{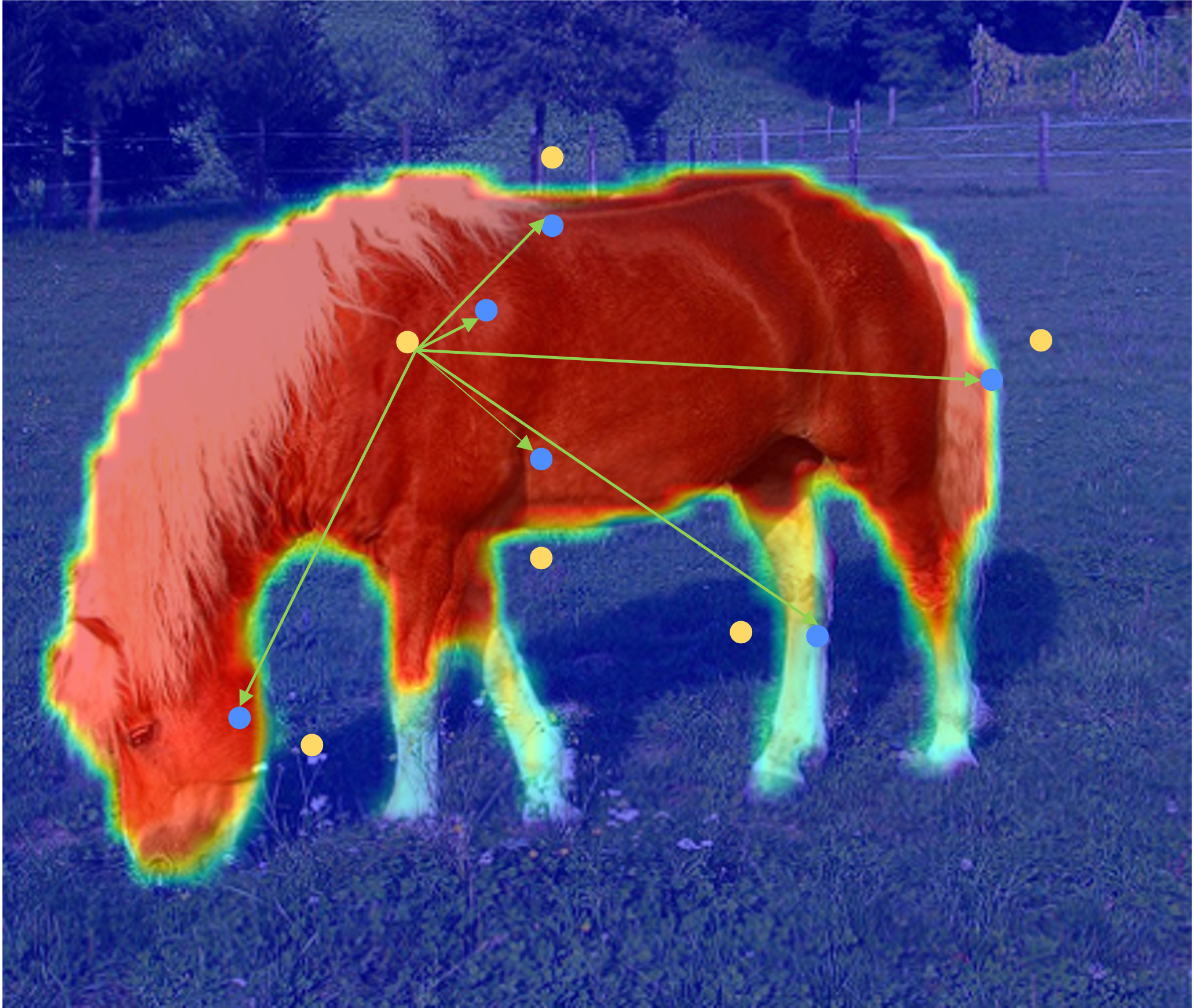} & 
\includegraphics[width=.47\linewidth, height=.4\linewidth]{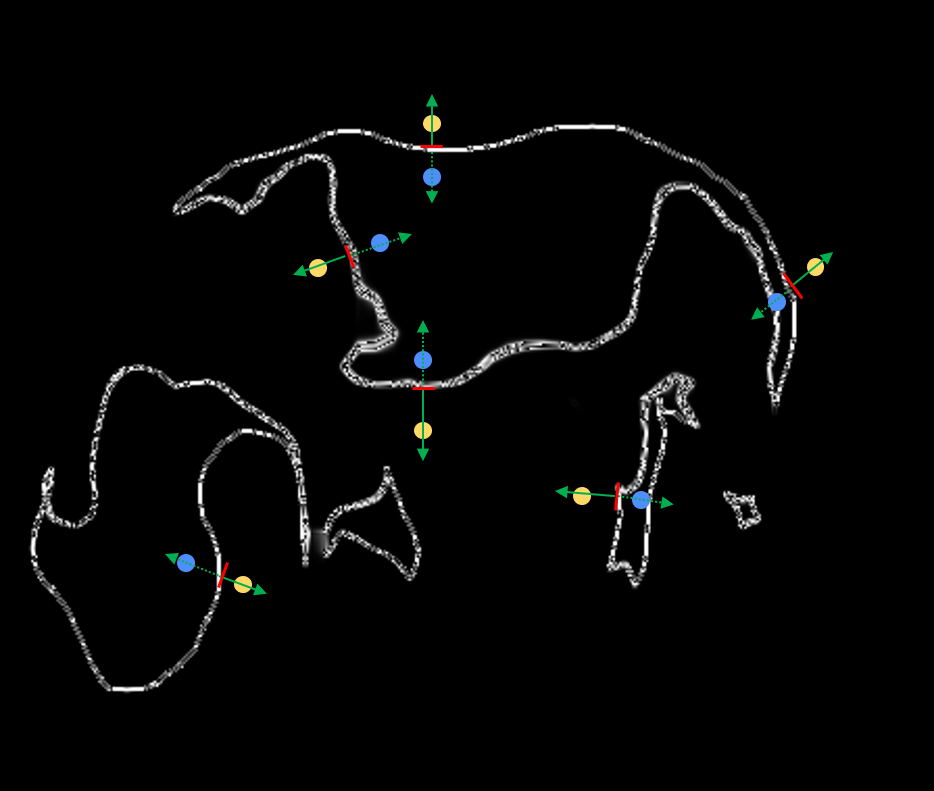}\\
(c) & (d)
\end{tabular}
\caption{Illustration of inward/outward point set division. (a) original image; (b) segmentation map during training; (c) pseudo mask; (d) in/out-ward division based on boundary map.}
\label{fig:beacon illustration}
\end{figure}

\begin{algorithm}[b]
\SetAlgoLined
\SetKwData{Orient}{orient}\SetKwData{Dense}{dense}\SetKwData{Mask}{mask}\SetKwData{PosId}{PosId}
    
\SetKwFunction{InOutDiv}{InOutDiv}\SetKwFunction{Sim}{$f$}\SetKwFunction{sg}{sg}\SetKwFunction{Sign}{Sign}
\SetKwInOut{Input}{Input}
\SetKwInOut{Output}{Output}
\Input{orientation map $M$; dense feature map $\Tilde{y}$; soft pseudo mask $y$; parameters $steps$, $k$
}
\Output{BEACON loss $\mathcal{L}_{beacon}$}

\nosemic
\textcolor{blue}{// select in/out-ward point sets from $M$:}\;
\dosemic
$\Phi,~\Psi\leftarrow$\InOutDiv{$M,~steps$}\;
randomly select $k$ samples from $\Phi$, $\Psi$ as $I$, $O$\;
\nosemic
\textcolor{blue}{// anchor $I$, $O$ back onto $y$ and $\tilde{y}$ to yield inward set ($I^{d}, I^{m}$) and outward set ($O^{d}, O^{m}$) on dense feature map and pseudo mask:}\;
\dosemic
$I^{m},~O^{m}\leftarrow y^{I},~y^{O}$\;
$I^{d},~O^{d}\leftarrow \Tilde{y}^{I},~\Tilde{y}^{O}$\;
\nosemic
\textcolor{blue}{// calculate similarity matrix $S$ for the two sets:}\;
\dosemic$S^{d}\leftarrow$\Sim{\sg{$I^{d}$},$~O^{d}$}\;
$S^{m}\leftarrow$\Sim{$I^{m}$,$~O^{m}$}\;
\nosemic
\textcolor{blue}{// obtain point-wise signs:}\;
\dosemic $sign^{I}, sign^{O}\leftarrow$\Sign{$S_{i,o}^{m},~S_{i,o}^{d}$}\;
\nosemic
\textcolor{blue}{// calculate and return loss:}\;
$\mathcal{L}_{beacon}\leftarrow \mathop\frac{1}{\vert{O}\vert}\sum\limits_{o \in O}\log(sign^{O}\cdot \mathop\frac{1}{\vert{I}\vert}\sum\limits_{i\in I}S^{d}_{i,o})$\;
\dosemic
\pushline\pushline\pushline$+\mathop\frac{1}{\vert{I}\vert}\sum\limits_{i \in I}\log(sign^{I}\cdot \mathop\frac{1}{\vert{O}\vert}\sum\limits_{o\in O}S^{d}_{i,o})$\;
 
\caption{BEACON algorithm. $\text{InOutDiv}(\cdot)$ is described in the Supplemental Material.}
\label{ag:beacon}
\end{algorithm}

When the decoder of a typical WSSS network is trained, pseudo labels generated from the encoder output are used to supervise the learning process. Consequently, these pseudo labels are key to the final segmentation performance. Although some recent work improves the quality of pseudo labels by introducing an extra processing stage, such as an AffinityNet~\cite{ahn2018affinitynet} or conditional random fields~\cite{lafferty2001conditional}, to enhance implicit boundary smoothness, the resulting hard masks lead to supervision bias during training. To alleviate this and achieve more consistent segmentation results across object boundaries, we propose a novel boundary contrastive loss term, named Boundary Enhancement viA Contrastive Orientation Navigation (BEACON), to further improve our segmentation network. The detailed algorithm for BEACON is given in Algorithm~\ref{ag:beacon}.
    
We first form two boundary candidate point sets, an inward point set and an outward point set. To do so, we apply the Sobel operator~\cite{sobeldetector} on the segmentation map, and identify object boundary points as those points that exhibit the top 20\% largest gradient magnitudes. This allows to build reliable sets to select paired samples for contrastive learning. Note that the obtained boundary map (see Figure~\ref{fig:beacon illustration}(d) for an example) has a strong semantic meaning and is different from applying the Sobel operator directly on the input image. We obtain the gradient directions of the boundary points and quantise them to 8 directions corresponding to the 8-neighborhood of a pixel. Based on a step parameter, we then calculate a displacement from a boundary point along the gradient direction as well as the opposite direction to generate candidate points for the two sets.
 
Having obtained the inward and outward boundary point sets, we use the soft pseudo masks generated from our MCL encoder (Figure~\ref{fig:beacon illustration}(c)) and the dense map from the decoder to define a boundary contrastive loss term. As illustrated in Figure~\ref{fig:beacon illustration}(b), the segmentation map is far from perfect at the early training stages. Thus, we calculate the point-wise one-to-all similarity between the two sets on both the dense feature map and the pseudo mask to enhance the object boundary feature consistency. In particular, we define a sign function $\text{Sign}(\cdot)$ to identify if the similarity values calculated from the dense feature map $\tilde{y}$ coincide with those from the soft pseudo mask $y$ by comparing their scores to a threshold $\tau$.

\begin{algorithm}[t!]
\SetAlgoLined
\SetKwFunction{AND}{AND}
\SetKwInOut{Input}{Input}
\SetKwInOut{Output}{Output}
\Input{mask similarity matrix $S_{i,o}^{m}$; dense feature similarity matrix $S_{i,o}^{d}$}
\Output{point-wise signs $sign^{I}, sign^{O}$}

$S^{m}_{I}\leftarrow\mathop\frac{1}{\vert{O}\vert}\sum\limits_{o\in O}{S_{i,o}^{m}}$\;
$S^{d}_{I}\leftarrow\mathop\frac{1}{\vert{O}\vert}\sum\limits_{o\in O}{S_{i,o}^{d}}$\;

$FP^{I}\leftarrow$\AND{$\mathbb{I}(S^{m}_{I}>\tau)$, $\mathbb{I}(S^{d}_{I}<\tau)$}\;
$FN^{I}\leftarrow$\AND{$\mathbb{I}(S^{m}_{I}<\tau)$, $\mathbb{I}(S^{d}_{I}>\tau)$}\;
$TP^{I}\leftarrow$\AND{$\mathbb{I}(S^{m}_{I}<\tau)$, $\mathbb{I}(S^{d}_{I}<\tau)$}\;
$TN^{I}\leftarrow$\AND{$\mathbb{I}(S^{m}_{I}>\tau)$, $\mathbb{I}(S^{d}_{I}>\tau)$}\;
        
convert truth values of $FP^{I}$, $FN^{I}$, $TP^{I}$, and $TN^{I}$ into binary values \{-1, 1\}\;
\nosemic
\textcolor{blue}{// assign negative to actual condition negative cases:}\;
\dosemic
$TN^{I}\leftarrow -TN^{I}$\;
$FP^{I}\leftarrow -FP^{I}$\;
\nosemic
\textcolor{blue}{// compute signs for inward set:}\;
\dosemic
$sign^{I}\leftarrow  FN^{I} \cup TP^{I} \cup TN^{I} \cup FP^{I}$\;
\nosemic
\textcolor{blue}{// compute signs for outward set:}\;
\dosemic
$sign^{O}\leftarrow  FN^{O} \cup TP^{O} \cup TN^{O} \cup FP^{O}$\;

\caption{Sign$(\cdot)$ function.}
\label{ag:sign}
\end{algorithm}

The sign determines the direction of optimisation imposes on similarity as shown in Algorithm~\ref{ag:sign}. Intuitively, if $S^{d}<\tau$, the query in-out pair is recognised as dissimilar and thus a positive edge (P) is assigned. Furthermore, if $S^{m}<\tau$ is also satisfied, a true positive (TP) case is identified, yielding a similarity suppression (positive sign) to make them more dissimilar. Integrating TP, FP, FN, and TN cases, point-wise signs are obtained and are used to calculate the boundary contrastive loss.

The loss function for this training stage is expressed as
\begin{equation}
\label{eq:loss_seg}
\mathcal{L}_{seg}=\mathcal{L}_{ce}+\lambda\mathcal{L}_{beacon},
\end{equation}
where parameter $\lambda$ allows for balancing between the global pixel-wise cross entropy loss $\mathcal{L}_{ce}$ and the near-boundary pixel representation enhancement.

\section{Experimental Results}
\label{experiment}
\subsection{Implementation details}
\label{implementation}
Our experiments are conducted on the PASCAL VOC 2012 dataset~\cite{pascal-voc-2012} with 20 foreground classes and 1 background class. Following~\cite{hariharan2011sbd, wang2020seam}, we build an augmented training set with 10,582 images. During classification training, only the 20 foreground class logits are taken into consideration, while the background class activation map is estimated for pairwise regional contrast. We use cosine similarity to compute the pairwise cost matrix and use Sinkhorn iteration~\cite{cuturi2013sinkhorn} for fast computation of the EMD.
    
Our MuSCLe implementation comprises an EfficientNet encoder~\cite{tan2019efficientnet} and a BiFPN decoder~\cite{effdet}. To efficiently scale up the model, we use batch sizes of 16, 8 and 6 for EfficientNet-b3, EfficientNet-b5, and EfficientNet-b7, respectively, with the same decoder which has 3 BiFPN layers. Experiments are conducted on an RTX 3090 GPU using PyTorch~\cite{NEURIPS2019PyTorch}. The input of image level contrast and classification head is resized while keeping the original image aspect ratio and padded to $448\times448$, while the pixel level contrast and pairwise regional contrast heads use random crops of size $224\times224$ as inputs. CRF and affinity refinement~\cite{ahn2018affinitynet} are executed after SAM output to generate pseudo masks. 

\subsection{Improved CAM quality}
We quantitatively evaluate the effectiveness of each component of our MuSCLe approach in Table~\ref{tab:ablation_components}. From there, it is evident that each proposed module leads to a notable performance increase. Following common practice~\cite{ahn2018affinitynet,wang2020seam, chang2020subcat}, test time augmentation (TTA) with multi-scale inference gives a further improvement of 2.5\%-3\%. Compared to ordinary CAM methods, our multi-contrast learning encoder improves CAM quality by a large margin (+6.8\%).
    
\begin{table}[b!]
\centering
\small
\begin{tabular}{cccccc}
\hline
\multirow{2}{*}{\textbf{HCL}}& \multirow{2}{*}{\textbf{IMC}} & \multirow{2}{*}{\textbf{PIXC}} & \multirow{2}{*}{\textbf{PRC}} & \textbf{single scale} & \textbf{multi-scale} \\
&  &  &  & \textbf{mIoU} & \textbf{mIoU} \\
\hline
& & & & 48.5 & 51.6\\
\checkmark & & & & 53.3 & 55.7 \\
\checkmark & \checkmark & & & 54.3 & 57.2 \\
\checkmark & \checkmark & \checkmark & & 54.8 & 57.6 \\
\checkmark & \checkmark & \checkmark & \checkmark & \textbf{55.3} & \textbf{58.4} \\
\hline
\end{tabular}
\caption{Ablation study for MCL encoder. mIoU (\%) reflects pseudo CAM quality on \textit{train} set. HCL=hybrid classification loss; IMC=image level contrast; PIXC=pixel level contrast; PRC=pairwise regional contrast.}
\label{tab:ablation_components}
\end{table}

Table~\ref{tab:pseudo_label_quality} compares the pseudo label quality of our method with other state-of-the-art (SOTA) approaches. As we can see, our MCL clearly outperforms the other methods, improving the CAM of AffinityNet~\cite{ahn2018affinitynet} by 10.4\% and the result of SEAM~\cite{wang2020seam} by 3.0\%. Although the improvement with affinity refinement is less pronounced compared to that for raw CAMs, we still obtain the highest mIoU of 64.6\%. We conjecture that this is because our CAM trained by MCL is denser and more continuous, and thus the affinity transformation barely enhances local feature representation with adjacent context.
    
\begin{table}[t!]
\centering
\small
\begin{tabular}{lcc}
\hline
\textbf{method} & \textbf{CAM} & \textbf{CAM+RW} \\
\hline
AffinityNet~\cite{ahn2018affinitynet} & 48.0 & 58.1\\
IRN~\cite{ahn2019weakly} & 48.3 & 59.3 \\
SC-CAM~\cite{chang2020subcat} & 50.9 & 63.4 \\
SEAM~\cite{wang2020seam} & 55.4 & 63.6\\
MCL & \textbf{58.4} & \textbf{64.6} \\
\hline
\end{tabular}
\caption{Comparison of pseudo label mIoU results on \textit{train} set. RW=random walk with affinity transformation.}
\label{tab:pseudo_label_quality}
\end{table}
    
\begin{figure}[t!]
\centering
\includegraphics[width=.49\linewidth]{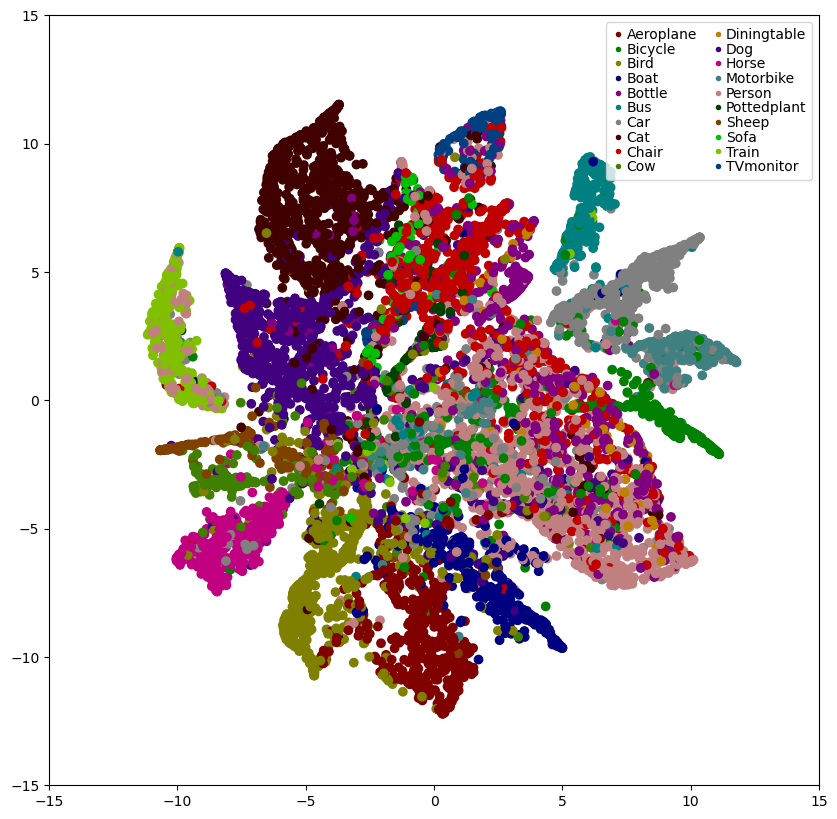} \hfill
\includegraphics[width=.49\linewidth]{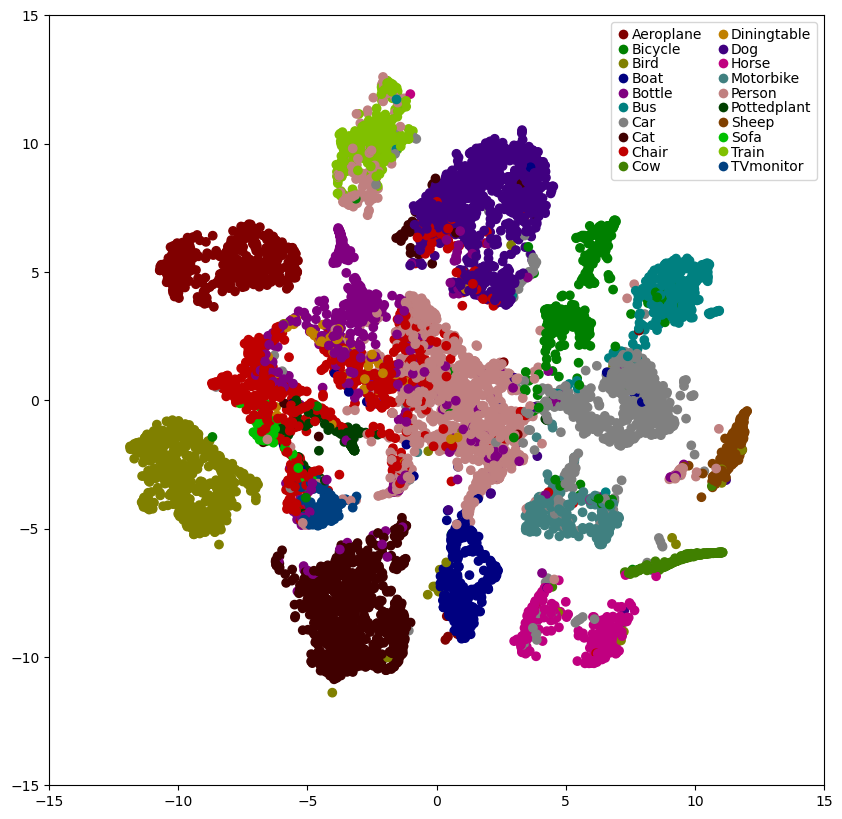}
\caption{t-SNE visualisations of the learned representations of encoder for SEAM (left) and our MCL (right).}
\label{fig:t-sne}
\end{figure}

In addition, we have visualised the learned representations from SEAM and our MCL using t-SNE dimensionality reduction~\cite{van2008visualizing}. As can be seen from the obtained results in Figure~\ref{fig:t-sne}, the classes are better separated in the MCL visualisation, while for SEAM we can observe significant overlap between classes and higher variation within each class.
    
\begin{table}[b!]
\centering
\small
\begin{tabular}{lccc}
\hline
\textbf{backbone} & \textbf{\# BiFPN} & \textbf{BEACON} & \textbf{mIoU}\\
\hline
EfficientNet-b3 & 1 & no & 60.7\\
EfficientNet-b3 & 2 & no & 61.3\\
EfficientNet-b3 & 3 & no & 63.2 \\ 
EfficientNet-b5 & 3 & no & 63.8 \\
EfficientNet-b7 & 3 & no & 64.5\\
EfficientNet-b3 & 3 & yes & 64.1 \\
EfficientNet-b5 & 3 & yes & 65.2 \\
EfficientNet-b7 & 3 & yes & \textbf{66.1}\\
\hline
\end{tabular}
\caption{Ablation study for segmentation network architecture. mIoU reflects segmentation performance on \textit{val} set.}
\label{tab:ablation_segmentation}
\end{table}

\begin{table}[t!]
\centering
\small
\begin{tabular}{cccccc}
\hline
\multirow{2}{*}{\textbf{$\lambda$}}& \multirow{2}{*}{\textbf{steps}}&  \multirow{2}{*}{\textbf{$k$}}& \multirow{2}{*}{\textbf{$\tau$}}& \textbf{single scale} & \textbf{multi scale}\\
& & &  & \textbf{mIoU} & \textbf{mIoU}\\
\hline
0 & n/a & n/a & n/a & 60.3~\vline~60.5 & 63.8~\vline~64.1 \\
0.05 & 7 & 128 & 0.5 & 60.5~\vline~60.4 & 64.1~\vline~63.8 \\
0.05 & 7 & 64 & $\mu_{m}$ & 60.1~\vline~60.6 & 63.8~\vline~63.9\\
0.05 & 7 & 128 & $\mu_{m}$ & \textbf{61.7}~\vline~61.6 & 65.2~\vline~\textbf{66.1}\\
0.05 & 7 & 256 & $\mu_{m}$ & 60.7~\vline~61.5 & 64.4~\vline~64.6 \\
0.05 & 5 & 128 & $\mu_{m}$ & 60.0~\vline~61.7 & 64.5~\vline~65.1\\
0.05 & 9 & 128 & $\mu_{m}$ & 59.7~\vline~61.1 & 63.9~\vline~64.2\\
0.1 & 7 & 128 & $\mu_{m}$ & 60.4~\vline~60.9 & 64.7~\vline~64.3\\
\hline
\end{tabular}
\caption{BEACON ablation study. mIoU reflects the segmentation performance on \textit{val} set. $\mu_{m}$ denotes the mean of similarity matrix derived from the soft mask. In the result columns, the left and right values denote MuSCLe-b5 and MuSCLe-b7 results, respectively.}
\label{tab:ablation_beacon}
\end{table}

\begin{table}[b!]
\centering
\small
\setlength{\tabcolsep}{0.2em}
\begin{tabular}{lcccc}
\hline
\textbf{method}& \textbf{parameters}  & \textbf{label} & \textbf{validation} & \textbf{test}\\
& \textbf{$[10^{6}]$} &  & \textbf{mIoU} & \textbf{mIoU}\\
\hline
AffinityNet\textsubscript{CVPR2018}~\cite{ahn2018affinitynet} & $268.7$ & I & 61.7 & 63.7 \\
FickleNet\textsubscript{CVPR2019}~\cite{lee2019ficklenet} & $306.4$ & I+S & 64.9 & 65.3 \\
OAA\textsubscript{ICCV2019}~\cite{jiang2019oaa} & $172.8$ & I+S & 65.2 & 66.4 \\
SEAM\textsubscript{CVPR2020}~\cite{wang2020seam} & $218.8$ & I & 64.5 & 65.7 \\
SC-CAM\textsubscript{CVPR2020}~\cite{chang2020subcat} & $204.5$ & I & \textbf{66.1} & 65.9 \\
BES\textsubscript{ECCV2020}~\cite{chen2020weakly} & $105.0$ & I & 65.7 & 66.6 \\
LayerCAM\textsubscript{TIP2021}~\cite{jiang2021layercam} & $218.4$ & I & 63.0 & 64.5 \\
\hline
MuSCLe-b5  & 44.9 & I & 65.2 & 66.7 \\
MuSCLe-b7  & 81.7 & I & \textbf{66.1} &  \textbf{67.3}\\
\hline
\end{tabular}
\caption{Comparison with SOTA WSSS methods in terms of mIoU on VOC2012 \textit{val} and \textit{test} set. I=image level label; I+S=image level label + saliency map. Parameters are counted on classification network and segmentation network together.}
\label{tab:SOTA}
\end{table}

\begin{table*}[t!]
\centering
\small
\renewcommand*{\arraystretch}{1.0}
\setlength{\tabcolsep}{0.2em}
\begin{tabular}{l|cccccccccccccccccccccc}
\hline
\textbf{method}& \textbf{\rotatebox{45}{bkg}} & \textbf{\rotatebox{45}{aero}} & \textbf{\rotatebox{45}{bike}} & \textbf{\rotatebox{45}{bird}} &  \textbf{\rotatebox{45}{boat}} & \textbf{\rotatebox{45}{bottle}} & \textbf{\rotatebox{45}{bus}} & \textbf{\rotatebox{45}{car}} & \textbf{\rotatebox{45}{cat}} & \textbf{\rotatebox{45}{chair}} & \textbf{\rotatebox{45}{cow}} & \textbf{\rotatebox{45}{table}} & \textbf{\rotatebox{45}{dog}} & \textbf{\rotatebox{45}{horse}} & \textbf{\rotatebox{45}{mbk}} & \textbf{\rotatebox{45}{person\ }} & \textbf{\rotatebox{45}{plant}} & \textbf{\rotatebox{45}{sheep}} & \textbf{\rotatebox{45}{sofa}} & \textbf{\rotatebox{45}{train}} & \textbf{\rotatebox{45}{tv}}\\
\hline
AffinityNet~\cite{ahn2018affinitynet} & 88.2 & 68.2 & 30.6 & 81.1 & 49.6 & 61.0 & 77.8 & 66.1 & 75.1 & 29.0 & 66.0 & 40.2 & 80.4 & 62.0 & 70.4 & 73.7 & 42.5 & 70.7 & 42.6 & \textbf{68.1} & 51.6 \\ 
               
FickleNet~\cite{lee2019ficklenet} & \textbf{89.5} & \textbf{76.6} & 32.6 & 74.6 & 51.5 & 71.7 & 83.4 & 74.4 & 83.6 & 24.1 & 73.4 & 47.4 & 78.2 & 74.0 & 68.8 & 73.2 & 47.8 & 79.9 & 37.0 & 57.3 & \textbf{64.6} \\ 

SEAM~\cite{wang2020seam} & 88.8 & 68.5 & \textbf{33.3} & \textbf{85.7} & 40.4 & 67.3 & 78.9 & 76.3 & 81.9 & 29.1 & 75.5 & 48.1 & 79.9 & 73.8 & 71.4 & \textbf{75.2} & 48.9 & 79.8 & 40.9 & 58.2 & 53.0 \\ 
                
SC-CAM~\cite{chang2020subcat}  & 88.8 & 51.6 & 30.3 & 82.9 & 53.0 & \textbf{75.8} & 88.6 & 74.8 & \textbf{86.6} & 32.4 & 79.9 & \textbf{53.8} & \textbf{82.3} & 78.5 & 70.4 & 71.2 & 40.2 & 78.3 & 42.9 & 66.8 & 58.8 \\ 
    
BES~\cite{chen2020weakly}  & 88.9 & 74.1 & 29.8 & 81.3 & \textbf{53.3} & 69.9 & \textbf{89.4} & \textbf{79.8} & 84.2 & 27.9 & 76.9 & 46.6 & 78.8 & 75.9 & 72.2 & 70.4 & \textbf{50.8} & 79.4 & 39.9 & 65.3 & 44.8 \\ 
    
MuSCLe-b7  & 88.4 & 75.5 & 31.8 & 76.3 & 50.3 & 71.7 & 85.8 & 77.4 & 72.1 & \textbf{33.4} & \textbf{90.2} & 49.1 & 75.6 & \textbf{82.8} & \textbf{73.6} & 70.4 & 49.0 & \textbf{85.9} & \textbf{47.7} & 65.3 & 36.1 \\ 
\hline
\end{tabular}
\caption{Category performance comparison on PASCAL VOC2012 \textit{val} set.}
\label{tab:CatResult}
\end{table*}

\subsection{Semantic segmentation training}
To investigate the impact of the decoder architecture on segmentation training using synthesised pseudo labels from the encoder, we test MuSCLe with different backbones, different numbers of BiFPN layers, and with and without BEACON. From Table~\ref{tab:ablation_segmentation}, we notice that BEACON leads to a consistent performance increase, while densifying BiFPN layers also gives notable improvement. In addition, scaling up the encoder backbone from b3 to b7 gives a 2.0\%/1.3\% boost with/without BEACON.
     
We perform a thorough ablation study, with results listed in Table~\ref{tab:ablation_beacon}, on our BEACON module to show the impact of different hyper-parameters and the effectiveness of BEACON. Since larger values of $\lambda$ in Eq.~(\ref{eq:loss_seg}) put more focus on near-boundary pixel enhancement and boundary map generation relies on accurate pixel-wise segmentation, as expected, too extreme $\lambda$ values do not lead to an improvement. For the step size walking towards the gradient orientation, we observe an optimal value of 7 with fewer steps not supporting sufficiently distinctive in/out-ward feature representation and more steps exceeding tiny object boundaries when selecting inward points along the inverse gradient orientation. Turning to the similarity threshold $\tau$, a halfway division of the similarity scores (\ie, $\tau=0.5$)  provides only a small improvement compared to a dynamic threshold, $\mu_{m}$, which is obtained as the mean of the similarity matrix derived from the soft mask. Selecting $k=128$ candidates of in-/outward pairs results in a good performance/efficiency trade-off. Overall, the best results are obtained by combining $\mu_{m}$ with $\lambda=0.05$, 7 steps, and $k=128$.

\subsection{Comparison with SOTA}
We compare MuSCLe with current SOTA methods in terms of performance, tuneable parameters and supervision in Table~\ref{tab:SOTA}. From there, we see that on the \textit{val} set we achieve performance on-par with SC-CAM~\cite{chang2020subcat} with merely 40\% of their learnable parameters, while outperforming all other methods. We also notice that even the smaller MuSCLe-b5 outperforms most other SOTA methods.

Importantly, we also observe MuSCLe to yield better performance on the \textit{test} set. With an mIoU of 66.7\% it is slightly superior to the previous best method of~\cite{chen2020weakly} and this with 57\% fewer parameters, based on an EfficientNet-b5 backbone. Further deepening the model to an EfficientNet-b7 structure, we obtain a new best mIoU of 67.3\% despite requiring fewer parameters than previous approaches and using only image level labels (in contrast to some other methods such as~\cite{lee2019ficklenet,jiang2019oaa} which rely on stronger supervision based on image labels in combination with saliency maps).

Looking at the class-wise performance on the \textit{val} set in Table~\ref{tab:CatResult}, MuSCLe gives the best result for 6 classes, more than any other method (SC-CAM is best for 4 categories). In particular, for the {\it cow}, {\it sheep}, and {\it sofa} classes, the performance is vastly superior to other approaches. On the other hand worse performance is obtained on the {\it tv} category. This is because we enforce contextual feature enhancement in our proposed method while TVs in the VOC dataset often appear together with other objects such as benches and tables. Further insight and examples are discussed in the Supplemental Material. 

\subsection{Qualitative segmentation results}
\begin{figure}[t!]
\centering
\renewcommand*{\arraystretch}{0.2}
\setlength{\tabcolsep}{0.03em}
\begin{tabular}{cccccc}
\includegraphics[height=1.3cm]{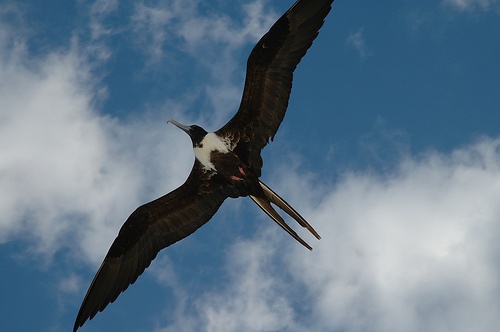} &
\includegraphics[height=1.3cm]{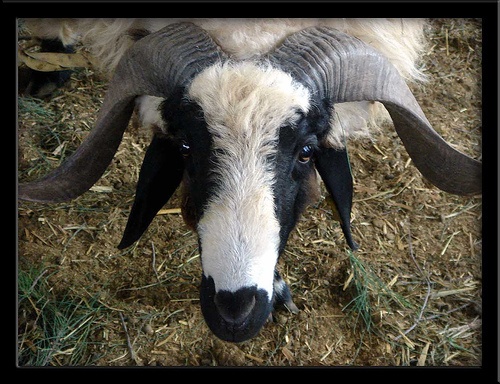} & 
\includegraphics[height=1.3cm]{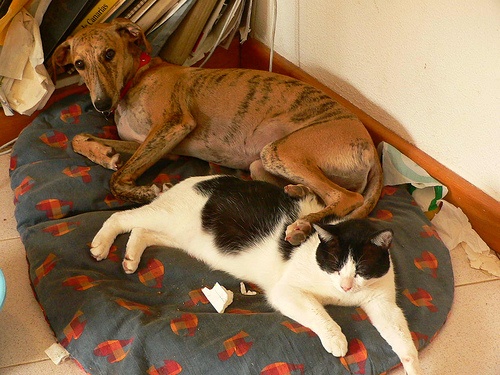} &
\includegraphics[height=1.3cm]{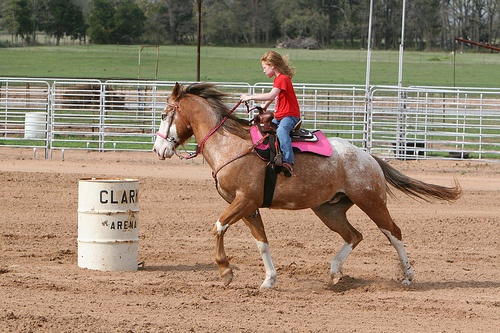} &
\includegraphics[height=1.3cm]{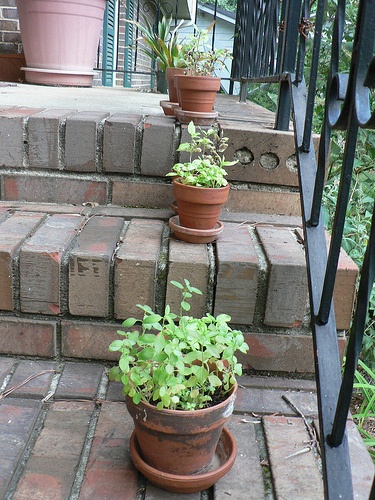}\\

\includegraphics[height=1.3cm]{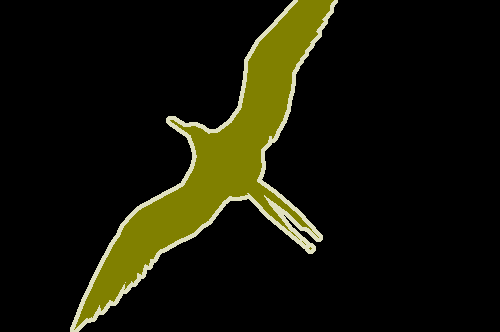} & 
\includegraphics[height=1.3cm]{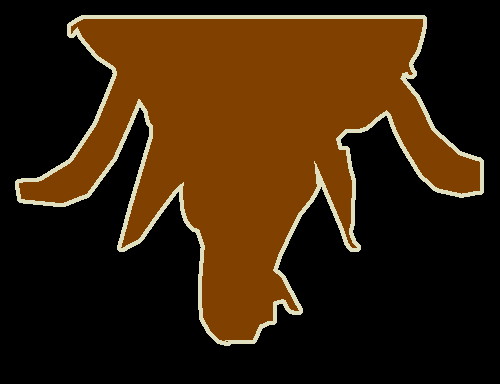} & 
\includegraphics[height=1.3cm]{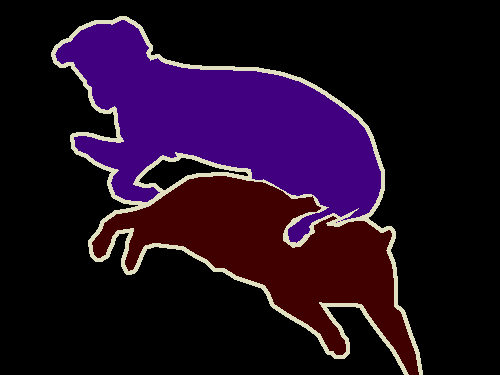} & 
\includegraphics[height=1.3cm]{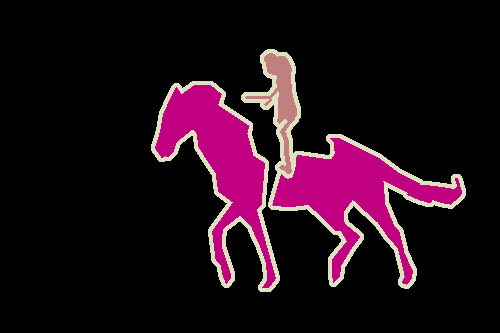} & 
\includegraphics[height=1.3cm]{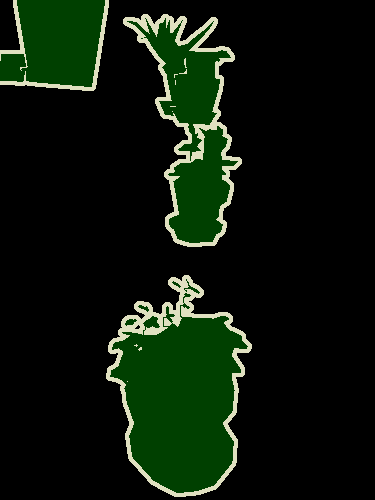} \\

\includegraphics[height=1.3cm]{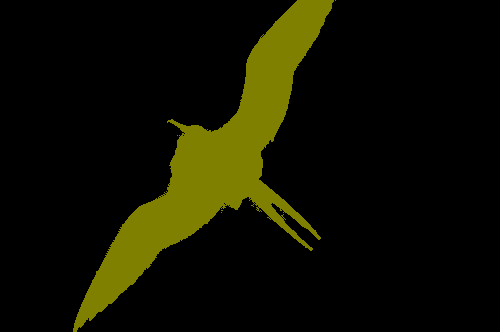} & 
\includegraphics[height=1.3cm]{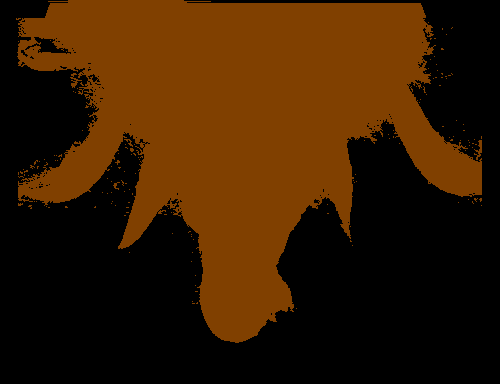} & 
\includegraphics[height=1.3cm]{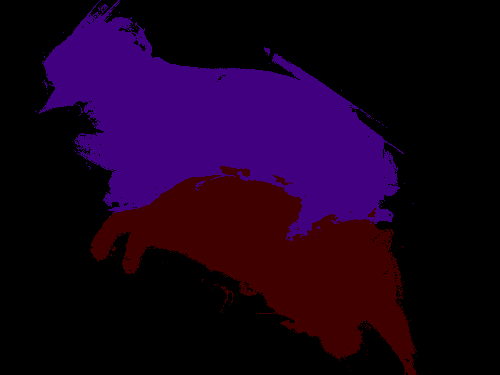} & 
\includegraphics[height=1.3cm]{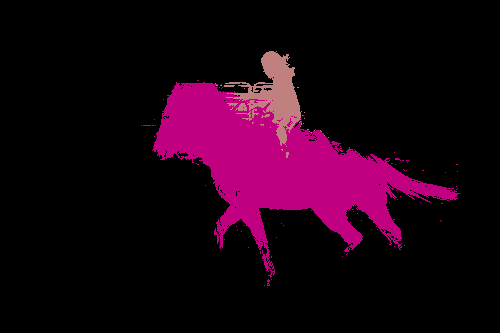} & 
\includegraphics[height=1.3cm]{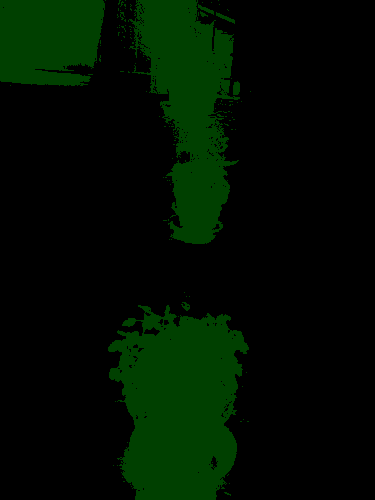} \\
\end{tabular}
\caption{Example segmentation results on PASCAL VOC2012 \textit{val} set. Top row: original image; middle row: ground truth; bottom row: segmentation result.}
\label{fig:seg_vis}
\vspace{-0.5cm}
\end{figure}
  
\begin{figure}[t!]
\centering
\renewcommand*{\arraystretch}{0.1}
\setlength{\tabcolsep}{0.02em}
\begin{tabular}{cccc}
\includegraphics[width=1.9cm]{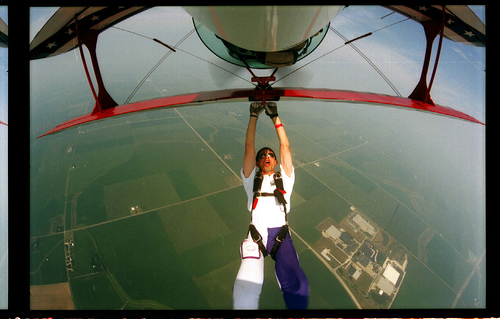} & 
\includegraphics[width=1.9cm]{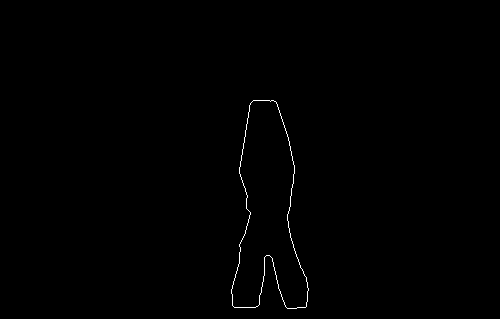} &  
\includegraphics[width=1.9cm]{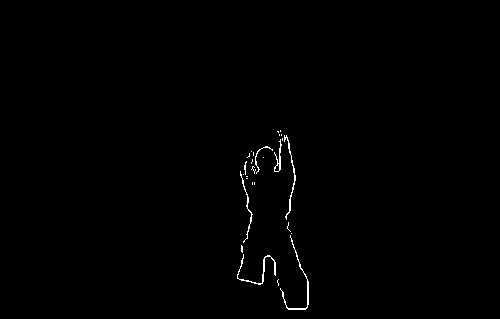} &
\includegraphics[width=1.9cm]{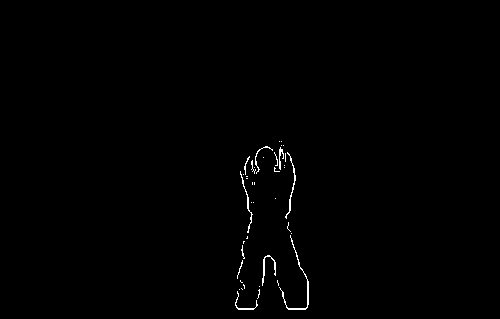} \\ 
    
\includegraphics[width=1.9cm]{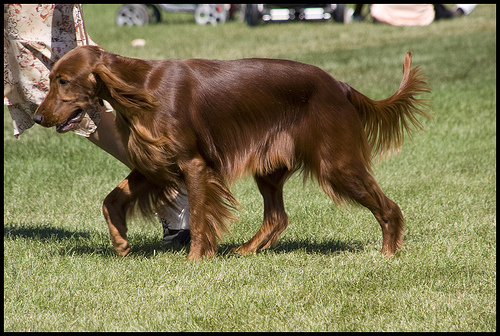} & 
\includegraphics[width=1.9cm]{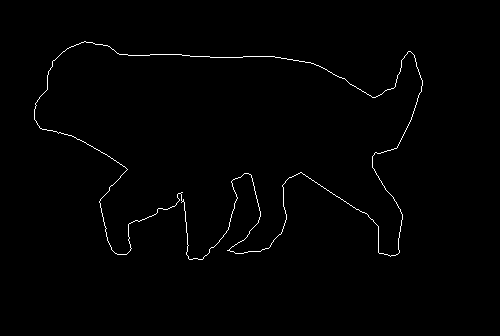} &  
\includegraphics[width=1.9cm]{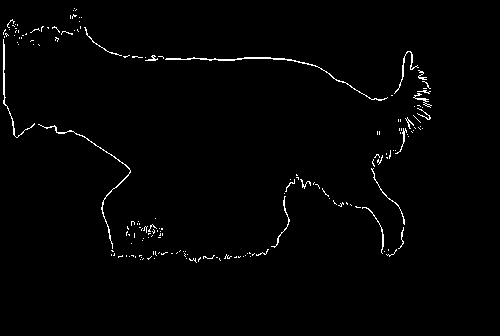} & 
\includegraphics[width=1.9cm]{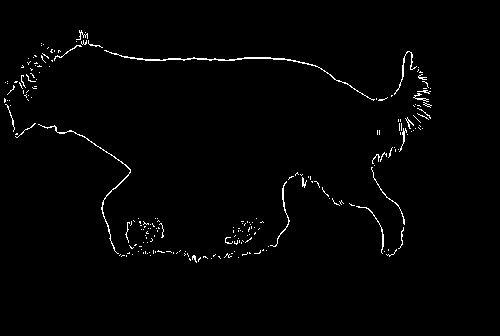}\\
    
\includegraphics[width=1.9cm]{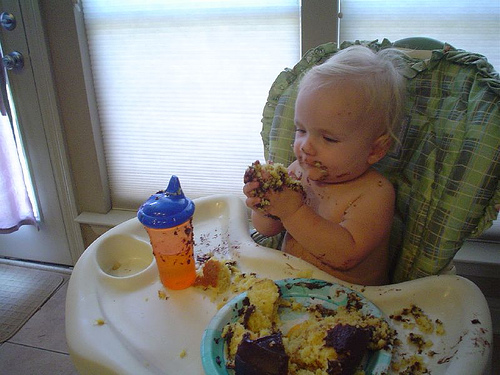} & 
\includegraphics[width=1.9cm]{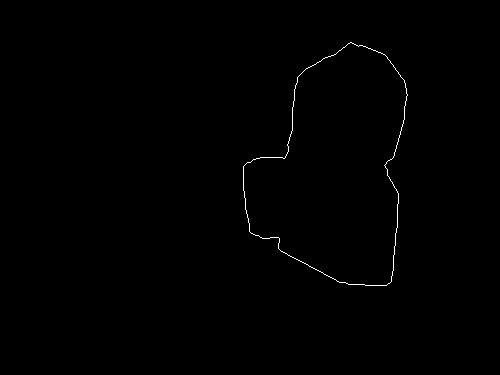} &  
\includegraphics[width=1.9cm]{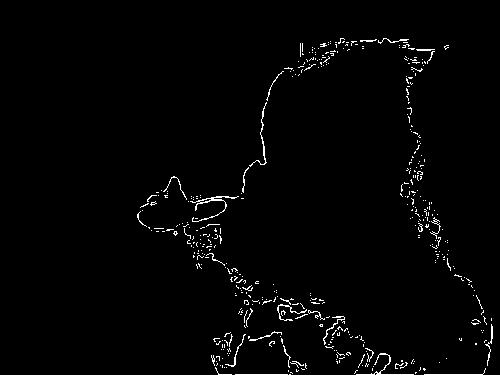} & 
\includegraphics[width=1.9cm]{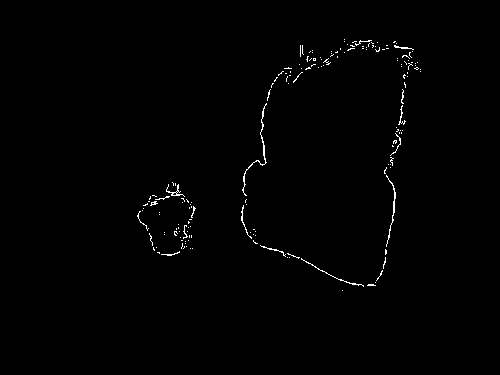}\\
    
\includegraphics[width=1.9cm]{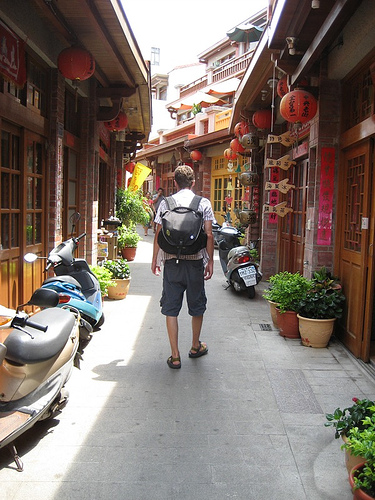} & 
\includegraphics[width=1.9cm]{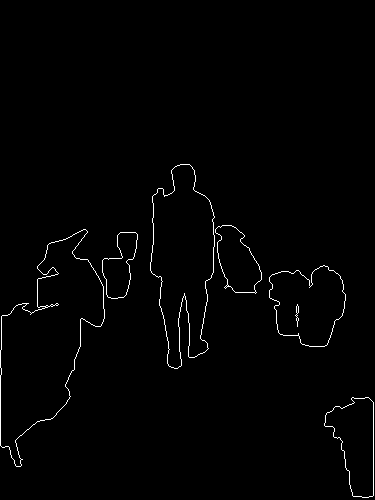} &  
\includegraphics[width=1.9cm]{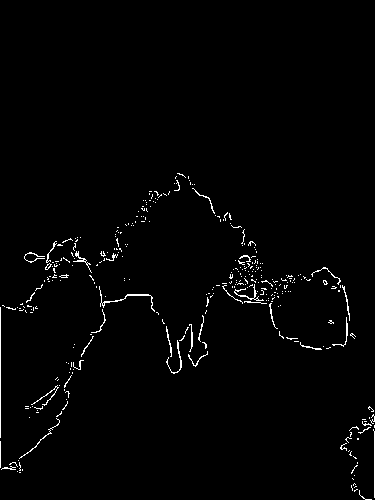} & 
\includegraphics[width=1.9cm]{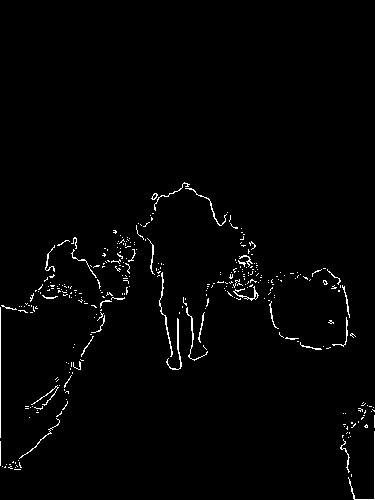}\\
\end{tabular}
\caption{Example boundary results on SBD \textit{trainval} set. From left to right: original image; class-agnostic semantic boundary label; semantic boundary without BEACON; semantic boundary with BEACON.}
\label{fig:sbd_vis}
\vspace{-0.5cm}
\end{figure}
  
We show some representative qualitative segmentation results obtained from MuSCLe-b7 in Figure~\ref{fig:seg_vis}, from which we can observe that detailed object boundaries are properly recovered (columns 1-2). For multi-label scenarios (columns 3-4), our model correctly distinguishes the  instances of each category, while multiple object instances at different scales and locations are also recognised (column 5), demonstrating the efficacy of the dynamic cropping and matching strategy. 

Figure~\ref{fig:sbd_vis} gives some typical samples from the SBD dataset~\cite{hariharan2011sbd}, illustrating the impact BEACON has on the obtained semantic boundaries. It is apparent that the semantic boundaries detected with BEACON are more complete and noise-robust compared to those without BEACON. 

\section{Conclusions}
In this paper, we exploit only image-level annotation to accomplish weakly supervised semantic segmentation. For this, we have proposed a novel MuSCLe framework which comprises an MCL encoder and a BEACON decoder. The former is designed to improve the initial CAM response via contrastive learning at different levels, while the latter aims to explicitly enhance feature representations around object boundaries through a contrastive scheme. Extensive experiments have demonstrated that, with significantly fewer parameters, MuSCLe achieves SOTA performance on the PASCAL VOC2012 dataset, while ablation studies and visualisations further illustrate the efficacy and efficiency of our proposed approach. Notably, this is achieved on a single GPU, unlike most existing work in the area. In future work, we will investigate extending our model to multi-GPU settings to further boost the performance.

{\small
\bibliographystyle{ieee_fullname}
\bibliography{egbib}

\begin{thebibliography}{10}\itemsep=-1pt

\bibitem{ahn2019weakly}
Jiwoon Ahn, Sunghyun Cho, and Suha Kwak.
\newblock Weakly supervised learning of instance segmentation with inter-pixel
  relations.
\newblock In {\em IEEE Conference on Computer Vision and Pattern Recognition},
  pages 2209--2218, 2019.

\bibitem{ahn2018affinitynet}
Jiwoon Ahn and Suha Kwak.
\newblock Learning pixel-level semantic affinity with image-level supervision
  for weakly supervised semantic segmentation.
\newblock In {\em IEEE Conference on Computer Vision and Pattern Recognition},
  pages 4981--4990, 2018.

\bibitem{bachman2019infomax}
Philip Bachman, R~Devon Hjelm, and William Buchwalter.
\newblock Learning representations by maximizing mutual information across
  views.
\newblock In {\em Neural Information Processing Systems}, pages 15535--15545,
  2019.

\bibitem{bearman2016points}
Amy Bearman, Olga Russakovsky, Vittorio Ferrari, and Li Fei-Fei.
\newblock What’s the point: Semantic segmentation with point supervision.
\newblock In {\em European Conference on Computer Vision}, pages 549--565.
  Springer, 2016.

\bibitem{cao2019gcnet}
Yue Cao, Jiarui Xu, Stephen Lin, Fangyun Wei, and Han Hu.
\newblock Gcnet: Non-local networks meet squeeze-excitation networks and
  beyond.
\newblock In {\em International Conference on Computer Vision Workshops}, pages
  0--0, 2019.

\bibitem{chang2020subcat}
Yu-Ting Chang, Qiaosong Wang, Wei-Chih Hung, Robinson Piramuthu, Yi-Hsuan Tsai,
  and Ming-Hsuan Yang.
\newblock Weakly-supervised semantic segmentation via sub-category exploration.
\newblock In {\em IEEE Conference on Computer Vision and Pattern Recognition},
  pages 8991--9000, 2020.

\bibitem{chen2020weakly}
Liyi Chen, Weiwei Wu, Chenchen Fu, Xiao Han, and Yuntao Zhang.
\newblock Weakly supervised semantic segmentation with boundary exploration.
\newblock In {\em European Conference on Computer Vision}, pages 347--362.
  Springer, 2020.

\bibitem{chen2020simclr}
Ting Chen, Simon Kornblith, Mohammad Norouzi, and Geoffrey Hinton.
\newblock A simple framework for contrastive learning of visual
  representations.
\newblock In {\em International Conference on Machine Learning}, pages
  1597--1607. PMLR, 2020.

\bibitem{chen2021simsam}
Xinlei Chen and Kaiming He.
\newblock Exploring simple siamese representation learning.
\newblock In {\em IEEE Conference on Computer Vision and Pattern Recognition},
  pages 15750--15758, 2021.

\bibitem{cuturi2013sinkhorn}
Marco Cuturi.
\newblock Sinkhorn distances: Lightspeed computation of optimal transport.
\newblock {\em Neural Information Processing Systems}, 26:2292--2300, 2013.

\bibitem{dai2015bbox2}
Jifeng Dai, Kaiming He, and Jian Sun.
\newblock Boxsup: Exploiting bounding boxes to supervise convolutional networks
  for semantic segmentation.
\newblock In {\em IEEE International Conference on Computer Vision}, pages
  1635--1643, 2015.

\bibitem{doersch2015position}
Carl Doersch, Abhinav Gupta, and Alexei~A Efros.
\newblock Unsupervised visual representation learning by context prediction.
\newblock In {\em IEEE International Conference on Computer Vision}, pages
  1422--1430, 2015.

\bibitem{pascal-voc-2012}
M. Everingham, L. Van~Gool, C.~K.~I. Williams, J. Winn, and A. Zisserman.
\newblock The {PASCAL} {V}isual {O}bject {C}lasses {C}hallenge 2012 {(VOC2012)}
  {R}esults.
\newblock
  http://www.pascal-network.org/challenges/VOC/voc2012/workshop/index.html.

\bibitem{autonomous_driving}
Di Feng, Christian Haase-Schütz, Lars Rosenbaum, Heinz Hertlein, Claudius
  Gläser, Fabian Timm, Werner Wiesbeck, and Klaus Dietmayer.
\newblock Deep multi-modal object detection and semantic segmentation for
  autonomous driving: Datasets, methods, and challenges.
\newblock {\em IEEE Transactions on Intelligent Transportation Systems},
  22(3):1341--1360, 2021.

\bibitem{grill2020byol}
Jean-Bastien Grill, Florian Strub, Florent Altch{\'e}, Corentin Tallec, Pierre
  Richemond, Elena Buchatskaya, Carl Doersch, Bernardo Avila~Pires, Zhaohan
  Guo, Mohammad Gheshlaghi~Azar, et~al.
\newblock Bootstrap your own latent-a new approach to self-supervised learning.
\newblock {\em Neural Information Processing Systems}, 33, 2020.

\bibitem{hariharan2011sbd}
Bharath Hariharan, Pablo Arbel{\'a}ez, Lubomir Bourdev, Subhransu Maji, and
  Jitendra Malik.
\newblock Semantic contours from inverse detectors.
\newblock In {\em IEEE International Conference on Computer Vision}, pages
  991--998. IEEE, 2011.

\bibitem{henaff2020cpcv2}
Olivier Henaff.
\newblock Data-efficient image recognition with contrastive predictive coding.
\newblock In {\em International Conference on Machine Learning}, pages
  4182--4192. PMLR, 2020.

\bibitem{sobeldetector}
FG Irwin et~al.
\newblock An isotropic 3x3 image gradient operator.
\newblock {\em Presentation at Stanford AI Project}, 2014(02), 1968.

\bibitem{jiang2019oaa}
Peng-Tao Jiang, Qibin Hou, Yang Cao, Ming-Ming Cheng, Yunchao Wei, and Hong-Kai
  Xiong.
\newblock Integral object mining via online attention accumulation.
\newblock In {\em IEEE International Conference on Computer Vision}, pages
  2070--2079, 2019.

\bibitem{jiang2021layercam}
Peng-Tao Jiang, Chang-Bin Zhang, Qibin Hou, Ming-Ming Cheng, and Yunchao Wei.
\newblock {LayerCAM}: Exploring hierarchical class activation maps for
  localization.
\newblock {\em IEEE Transactions on Image Processing}, 30:5875--5888, 2021.

\bibitem{khoreva2017bbox3}
Anna Khoreva, Rodrigo Benenson, Jan Hosang, Matthias Hein, and Bernt Schiele.
\newblock Simple does it: Weakly supervised instance and semantic segmentation.
\newblock In {\em IEEE Conference on Computer Vision and Pattern Recognition},
  pages 876--885, 2017.

\bibitem{khosla2020supervised}
Prannay Khosla, Piotr Teterwak, Chen Wang, Aaron Sarna, Yonglong Tian, Phillip
  Isola, Aaron Maschinot, Ce Liu, and Dilip Krishnan.
\newblock Supervised contrastive learning.
\newblock {\em arXiv preprint arXiv:2004.11362}, 2020.

\bibitem{kolesnikov2016sec}
Alexander Kolesnikov and Christoph~H Lampert.
\newblock Seed, expand and constrain: Three principles for weakly-supervised
  image segmentation.
\newblock In {\em European Conference on Computer Vision}, pages 695--711.
  Springer, 2016.

\bibitem{lafferty2001conditional}
John~D. Lafferty, Andrew McCallum, and Fernando Pereira.
\newblock Conditional {Random} {Fields}: Probabilistic models for segmenting
  and labeling sequence data.
\newblock In {\em International Conference on Machine Learning}, pages
  282--289, 2001.

\bibitem{lee2019ficklenet}
Jungbeom Lee, Eunji Kim, Sungmin Lee, Jangho Lee, and Sungroh Yoon.
\newblock Ficklenet: Weakly and semi-supervised semantic image segmentation
  using stochastic inference.
\newblock In {\em IEEE Conference on Computer Vision and Pattern Recognition},
  pages 5267--5276, 2019.

\bibitem{li2017pairloss}
Yuncheng Li, Yale Song, and Jiebo Luo.
\newblock Improving pairwise ranking for multi-label image classification.
\newblock In {\em IEEE Conference on Computer Vision and Pattern Recognition},
  pages 3617--3625, 2017.

\bibitem{lin2016scribblesup}
Di Lin, Jifeng Dai, Jiaya Jia, Kaiming He, and Jian Sun.
\newblock Scribblesup: Scribble-supervised convolutional networks for semantic
  segmentation.
\newblock In {\em IEEE Conference on Computer Vision and Pattern Recognition},
  pages 3159--3167, 2016.

\bibitem{lin2017focal}
Tsung-Yi Lin, Priya Goyal, Ross Girshick, Kaiming He, and Piotr Doll{\'a}r.
\newblock Focal loss for dense object detection.
\newblock In {\em IEEE International Conference on Computer Vision}, pages
  2980--2988, 2017.

\bibitem{liu2020selfemd}
Songtao Liu, Zeming Li, and Jian Sun.
\newblock {Self-EMD}: Self-supervised object detection without imagenet.
\newblock {\em arXiv preprint arXiv:2011.13677}, 2020.

\bibitem{nam2014large}
Jinseok Nam, Jungi Kim, Eneldo~Loza Menc{\'{i}}a, Iryna Gurevych, and Johannes
  F{\"u}rnkranz.
\newblock Large-scale multi-label text classification—revisiting neural
  networks.
\newblock In {\em European Conference on Machine Learning and Principles and
  Practice of Knowledge Discovery in Databases}, pages 437--452. Springer,
  2014.

\bibitem{noroozi2016jigsaw}
Mehdi Noroozi and Paolo Favaro.
\newblock Unsupervised learning of visual representations by solving jigsaw
  puzzles.
\newblock In {\em European Conference on Computer Vision}, pages 69--84.
  Springer, 2016.

\bibitem{oord2018cpc}
Aaron van~den Oord, Yazhe Li, and Oriol Vinyals.
\newblock Representation learning with contrastive predictive coding.
\newblock {\em arXiv preprint arXiv:1807.03748}, 2018.

\bibitem{papandreou1502bbox1}
G Papandreou, L-Ch Chen, K Murphy, and AL Yuille.
\newblock Weakly-and semi-supervised learning of a {DCNN} for semantic image
  segmentation.
\newblock {\em arXiv preprint arXiv:1502.02734}, 2015.

\bibitem{NEURIPS2019PyTorch}
Adam Paszke, Sam Gross, Francisco Massa, Adam Lerer, James Bradbury, Gregory
  Chanan, Trevor Killeen, Zeming Lin, Natalia Gimelshein, Luca Antiga, Alban
  Desmaison, Andreas Kopf, Edward Yang, Zachary DeVito, Martin Raison, Alykhan
  Tejani, Sasank Chilamkurthy, Benoit Steiner, Lu Fang, Junjie Bai, and Soumith
  Chintala.
\newblock {PyTorch}: An imperative style, high-performance deep learning
  library.
\newblock In H. Wallach, H. Larochelle, A. Beygelzimer, F. d\textquotesingle
  Alch\'{e}-Buc, E. Fox, and R. Garnett, editors, {\em Neural Information
  Processing Systems}, pages 8024--8035. Curran Associates, Inc., 2019.

\bibitem{pathak2016impaint}
Deepak Pathak, Philipp Krahenbuhl, Jeff Donahue, Trevor Darrell, and Alexei~A
  Efros.
\newblock Context encoders: Feature learning by inpainting.
\newblock In {\em IEEE Conference on Computer Vision and Pattern Recognition},
  pages 2536--2544, 2016.

\bibitem{precise_agriculture}
Diego~Inácio Patrício and Rafael Rieder.
\newblock Computer vision and artificial intelligence in precision agriculture
  for grain crops: A systematic review.
\newblock {\em Computers and Electronics in Agriculture}, 153:69--81, 2018.

\bibitem{ronneberger2015unet}
Olaf Ronneberger, Philipp Fischer, and Thomas Brox.
\newblock {U-Net}: Convolutional networks for biomedical image segmentation.
\newblock In {\em Medical Image Computing and Computer Assisted Intervention},
  pages 234--241. Springer, 2015.

\bibitem{selvaraju2017grad}
Ramprasaath~R Selvaraju, Michael Cogswell, Abhishek Das, Ramakrishna Vedantam,
  Devi Parikh, and Dhruv Batra.
\newblock Grad-cam: Visual explanations from deep networks via gradient-based
  localization.
\newblock In {\em IEEE International Conference on Computer Vision}, pages
  618--626, 2017.

\bibitem{tan2019efficientnet}
Mingxing Tan and Quoc Le.
\newblock Efficientnet: Rethinking model scaling for convolutional neural
  networks.
\newblock In {\em International Conference on Machine Learning}, pages
  6105--6114. PMLR, 2019.

\bibitem{effdet}
Mingxing Tan, Ruoming Pang, and Quoc~V Le.
\newblock Efficientdet: Scalable and efficient object detection.
\newblock In {\em IEEE Conference on Computer Vision and Pattern Recognition},
  pages 10781--10790, 2020.

\bibitem{van2008visualizing}
Laurens Van~der Maaten and Geoffrey Hinton.
\newblock Visualizing data using t-{SNE}.
\newblock {\em Journal of Machine Learning Research}, 9(11), 2008.

\bibitem{vernaza2017rw_scribble}
Paul Vernaza and Manmohan Chandraker.
\newblock Learning random-walk label propagation for weakly-supervised semantic
  segmentation.
\newblock In {\em IEEE Conference on Computer Vision and Pattern Recognition},
  pages 7158--7166, 2017.

\bibitem{wang2020seam}
Yude Wang, Jie Zhang, Meina Kan, Shiguang Shan, and Xilin Chen.
\newblock Self-supervised equivariant attention mechanism for weakly supervised
  semantic segmentation.
\newblock In {\em IEEE Conference on Computer Vision and Pattern Recognition},
  pages 12275--12284, 2020.

\bibitem{wei2017erase}
Yunchao Wei, Jiashi Feng, Xiaodan Liang, Ming-Ming Cheng, Yao Zhao, and
  Shuicheng Yan.
\newblock Object region mining with adversarial erasing: A simple
  classification to semantic segmentation approach.
\newblock In {\em IEEE Conference on Computer Vision and Pattern Recognition},
  pages 1568--1576, 2017.

\bibitem{xie2021pixpro}
Zhenda Xie, Yutong Lin, Zheng Zhang, Yue Cao, Stephen Lin, and Han Hu.
\newblock Propagate yourself: Exploring pixel-level consistency for
  unsupervised visual representation learning.
\newblock In {\em IEEE Conference on Computer Vision and Pattern Recognition},
  pages 16684--16693, 2021.

\bibitem{zhang2020deepemd}
Chi Zhang, Yujun Cai, Guosheng Lin, and Chunhua Shen.
\newblock {DeepEMD}: Few-shot image classification with differentiable earth
  mover's distance and structured classifiers.
\newblock In {\em IEEE Conference on Computer Vision and Pattern Recognition},
  pages 12203--12213, 2020.

\bibitem{zhou2016cam}
Bolei Zhou, Aditya Khosla, Agata Lapedriza, Aude Oliva, and Antonio Torralba.
\newblock Learning deep features for discriminative localization.
\newblock In {\em IEEE Conference on Computer Vision and Pattern Recognition},
  pages 2921--2929, 2016.

\bibitem{iclr21pseudo}
Yuliang Zou, Zizhao Zhang, Han Zhang, Chun-Liang Li, Xiao Bian, Jia-Bin Huang,
  and Tomas Pfister.
\newblock {PseudoSeg}: Designing pseudo labels for semantic segmentation.
\newblock In {\em International Conference on Learning Representations}, 2021.

\end{thebibliography}
}

\end{document}